\documentclass{article}

     \usepackage[final]{neurips_2025}

\usepackage[utf8]{inputenc} 
\usepackage[T1]{fontenc}    
\usepackage{hyperref}       
\usepackage{url}            
\usepackage{booktabs}       
\usepackage{amsfonts}       
\usepackage{nicefrac}       
\usepackage{microtype}      
\usepackage{xcolor}         
\usepackage{amsmath} 
\usepackage{amsthm}
\usepackage{graphicx}
\usepackage{subfigure}
\usepackage{wrapfig} 

\usepackage{booktabs} 
\usepackage{makecell} 
\usepackage{multirow}
\usepackage{siunitx}
\definecolor{mygray}{gray}{0.5}
\usepackage{soul}

\usepackage{algorithmic}
\usepackage[ruled,linesnumbered]{algorithm2e}
\theoremstyle{plain}
\newtheorem{theorem}{Theorem}  
\newtheorem{lemma}{Lemma}  

\theoremstyle{definition}
\newtheorem{remark}{Remark}  

\title{OmniFC: Rethinking Federated Clustering via Lossless and Secure Distance Reconstruction}

\author{Jie Yan\quad Jing Liu\quad Zhong-Yuan Zhang\thanks{Corresponding author.}
\vspace{0.2cm}\\
\vspace{0.2cm}
Central University of Finance and Economics\\
\small{jieyan@email.cufe.edu.cn,\, liu\_jing0623@163.com,\, zhyuanzh@gmail.com}
}
\begin{document}

\maketitle

\begin{abstract}

Federated clustering (FC) aims to discover global cluster structures across decentralized clients without sharing raw data, making privacy preservation a fundamental requirement. There are two critical challenges: (1) privacy leakage during collaboration, and (2) robustness degradation due to aggregation of proxy information from non-independent and identically distributed (Non-IID) local data, leading to inaccurate or inconsistent global clustering. Existing solutions typically rely on model-specific local proxies, which are sensitive to data heterogeneity and inherit inductive biases from their centralized counterparts, thus limiting robustness and generality. We propose Omni Federated Clustering (OmniFC), a unified and model-agnostic framework. Leveraging Lagrange coded computing, our method enables clients to share only encoded data, allowing exact reconstruction of the global distance matrix—a fundamental representation of sample relationships—without leaking private information, even under client collusion. This construction is naturally resilient to Non-IID data distributions. This approach decouples FC from model-specific proxies, providing a unified extension mechanism applicable to diverse centralized clustering methods. Theoretical analysis confirms both reconstruction fidelity and privacy guarantees, while comprehensive experiments demonstrate OmniFC's superior robustness, effectiveness, and generality across various benchmarks compared to state-of-the-art methods. Code will be released.
\end{abstract}



\section{Introduction}
Traditional clustering methods presuppose centralized access to the entire dataset, enabling the construction of global structures such as cluster centroids or kernel matrices. However, in federated settings characterized by data fragmentation across clients and privacy constraints, this assumption breaks down, precluding direct application.

\begin{figure}[!t]
\centering
\includegraphics[height = 6cm, width = 12cm]{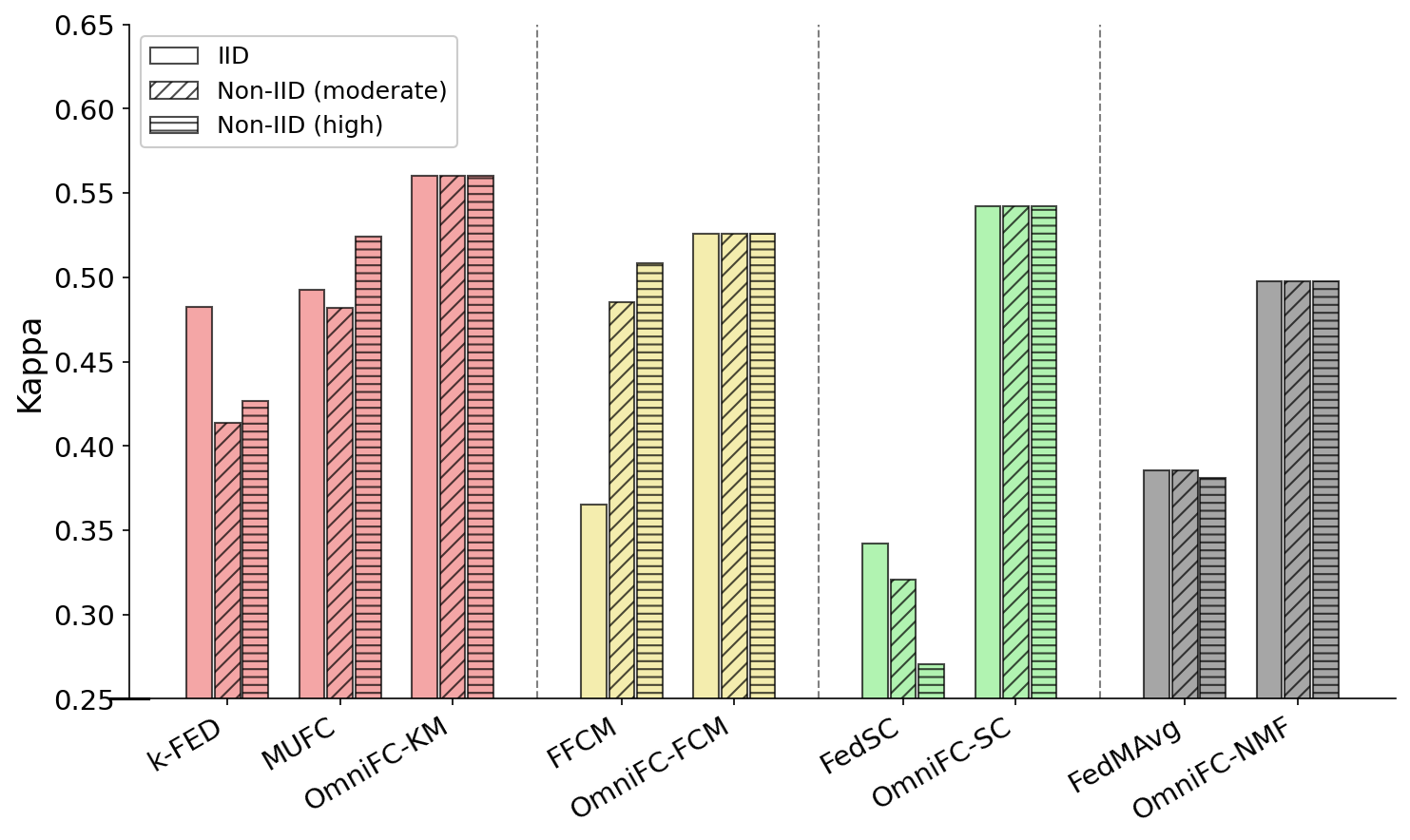}
\caption{\textbf{Robustness to heterogeneity.} We employ COIL-100 and 100 clients to compare the proposed OmniFC with the federated extensions of centralized clustering methods \cite{dennis2021heterogeneity, pan2023machine, stallmann2022towards, qiao2023federated, wang2022federated}. Compared to existing one-to-one extensions, OmniFC not only unifies the extension of centralized clustering methods but also achieves superior robustness and effectiveness.}
\label{case_noniid}
\end{figure}

To overcome this, federated clustering (FC) \cite{dennis2021heterogeneity} has emerged, enabling clients to collaboratively group data without sharing raw samples, and has found applications in client selection and exploratory data analysis. There are two fundamental challenges: (1) privacy leakage during collaboration, and (2) robustness degradation under non-independent and identically distributed (Non-IID) data. Existing FC methods approximate global structures by aggregating model-specific local proxies: federated k-means (KM) and fuzzy c-means (FCM) aggregate local cluster centroids \cite{dennis2021heterogeneity, stallmann2022towards, pan2023machine, xu2024jigsaw, yang2024greedy}, federated spectral clustering (SC) \cite{qiao2023federated} reconstructs the global kernel matrix from local low-rank factors, and federated non-negative matrix factorization (NMF) \cite{wang2022federated} aggregates local basis matrices. These proxies, however, are computed from biased client-specific datasets, fail to reliably capture global structures, leading to degraded robustness and performance (Fig. \ref{case_noniid}). Moreover, such methods are tightly bound to specific centralized clustering methods, inheriting restrictive inductive biases—e.g., data compactness in KM \cite{macqueen1967some} and FCM \cite{bezdek1984fcm}, data connectivity in SC \cite{ng2001spectral}, and low-rank representation in NMF \cite{lee2000algorithms}—thereby confining their performance to assumption-compliant data and limiting their generality.

This work addresses both limitations through a unifying perspective: reconstructing the global pairwise distance matrix, which offers a model-agnostic and fundamental representation of sample relationships, naturally resilient to the Non-IID problem. The key challenge, however, lies in securely computing this matrix without exposing private data. To this end, we propose \textit{Omni Federated Clustering (OmniFC)}, a novel framework that facilitates a unified extension from centralized clustering to FC through lossless and secure distance reconstruction. OmniFC comprises three main steps: local Lagrange-encoded sharing, global distance reconstruction, and cluster assignment. Each client initially encrypts its local data using Lagrange coded computing \cite{yu2019lagrange}, shares the encoded data with peers for pairwise distance computation, and subsequently transmits the resulting distances to the central server for constructing the global distance matrix. Finally, the global distance matrix can serve as input to centralized clustering methods for performing cluster assignment. Fig. \ref{case_noniid} demonstrates the superiority of OmniFC. With respect to distance reconstruction, the proposed OmniFC exhibits two salient features: 1) \textit{Efficacy}. Both theoretical and empirical analyses consistently demonstrate the capability for lossless reconstruction and robustness to the Non-IID problem. Benefiting from this, the proposed OmniFC achieves lossless federated extensions for pairwise-distance-dependent methods (e.g., SC) and enhances federated extensions for methods (e.g., KM) without explicit dependence on pairwise distances. 2) \textit{Security}. Theoretical analysis demonstrates that the privacy of local data is preserved during data sharing, as the encoded data prevents the inference of private information even under client collusion.  In summary, our contributions are threefold:

1) We propose OmniFC, a novel framework that facilitates a unified extension from centralized clustering to FC through lossless and secure distance reconstruction.

2) We establish theoretical assurances regarding the efficacy and security of distance reconstruction.

3) Experimental results show that our OmniFC outperforms SOTA methods on various benchmarks.

\section{Related Work}
\paragraph{Centralized Clustering.}
Traditional centralized clustering aggregates client-held local data on a central server for grouping, with methods making different assumptions—such as compactness \cite{macqueen1967some, ikotun2023k}, connectivity \cite{ng2001spectral, ding2024survey}, density \cite{ester1996density, xu2024scalable}, hierarchy \cite{gowda1978agglomerative, laber2024cohesion}, and low-rank representation \cite{lee2000algorithms, li2024tensorized} of the data distribution—to adapt to diverse datasets. However, these methods may become inapplicable due to privacy constraints that prevent the centralization of client data.

\paragraph{Federated Clustering (FC).}
Unlike centralized clustering, which requires collecting raw client data for model training, FC collects local proxies instead, thus strengthening user privacy protection. To handle this, several recent works have shifted from sharing local private data to exchanging local cluster centroids \cite{dennis2021heterogeneity, stallmann2022towards, xu2024jigsaw, yang2024greedy}, local basis matrices \cite{wang2022federated} or synthetic data \cite{yan2024sda}. Although these methods show promise, these methods either suffer from performance degradation caused by the Non-IID problem or achieve gains at the expense of privacy \cite{yan2024sda}.

\paragraph{Secure FC.}
Secure FC leverages advanced privacy-preserving techniques—including differential privacy \cite{dwork2014algorithmic}, machine unlearning \cite{cao2015towards}, and Lagrange coded computing \cite{yu2019lagrange}—to concurrently improve clustering efficacy and fortify data confidentiality. Existing methods typically focus on the effective and secure construction of either global cluster centroids for k-means \cite{li2022secure, pan2023machine, wang2024one, scott2025differentially, diaa2025fastlloyd} or a global kernel matrix for spectral clustering \cite{qiao2023federated}. Although promising, these methods remain limited by the Non-IID problem or fail to offer a model-agnostic solution. Moreover, they inherently retain assumptions—such as data compactness \cite{macqueen1967some, ikotun2023k} and connectivity \cite{ng2001spectral, ding2024survey}—from their centralized counterparts, limiting their effectiveness to compliant datasets and thereby reducing their practical applicability.

The most closely related work is SecFC \cite{li2022secure}, which also leverages Lagrange coded computing to improve clustering accuracy while preserving data confidentiality. In comparison, the proposed OmniFC exhibits the following distinctive characteristics: 1) \textit{Generality.} As a federated variant of k-means (KM), SecFC naturally inherits KM's assumption of data compactness, limiting their effectiveness to KM-friendly datasets. In contrast, the proposed unified and model-agnostic framework OmniFC accommodates diverse datasets by extending beyond KM to encompass alternatives such as spectral clustering and DBSCAN. 2) \textit{One-shot communication scheme.} To obtain more accurate cluster centroids, SecFC requires multiple rounds of communication between the clients and the server. However, multi-round training is unfeasible in some scenarios, like model markets, where users can solely purchase pre-trained models \cite{zhang2022dense}. In contrast, the proposed OmniFC, requiring merely one communication round, exhibits enhanced applicability in such scenarios.

\section{Omni Federated Clustering (OmniFC)}
This section begins with an overview of the problem definition and the OmniFC framework, followed by a detailed description of OmniFC, and concludes with its privacy analysis and complexity analysis.

\subsection{Overview} 

\paragraph{Problem Definition.}
Consider a real world dataset $\boldsymbol{X} \in \mathbb{R}^{n\times d}$ comprising $n$ $d$-dimensional samples $\{\boldsymbol{x}_i\}_{i = 1}^{n}$, which are distributed among $m$ clients, i.e., $\boldsymbol{X}=\bigcup_{j=1}^{m} \boldsymbol{X}_j$. FC aims to partition $\boldsymbol{X}$ into $k$ clusters with high intra-cluster similarity and low inter-cluster similarity while retaining $\boldsymbol{X}_j$ $(j \in [m] = \{1,\, 2,\, \cdots,\, m\})$ locally. A more detailed summary of notations is presented in Table \ref{notations} of the appendix.

\begin{figure}[!t]
\centering
\includegraphics[height = 5cm, width = 12cm]{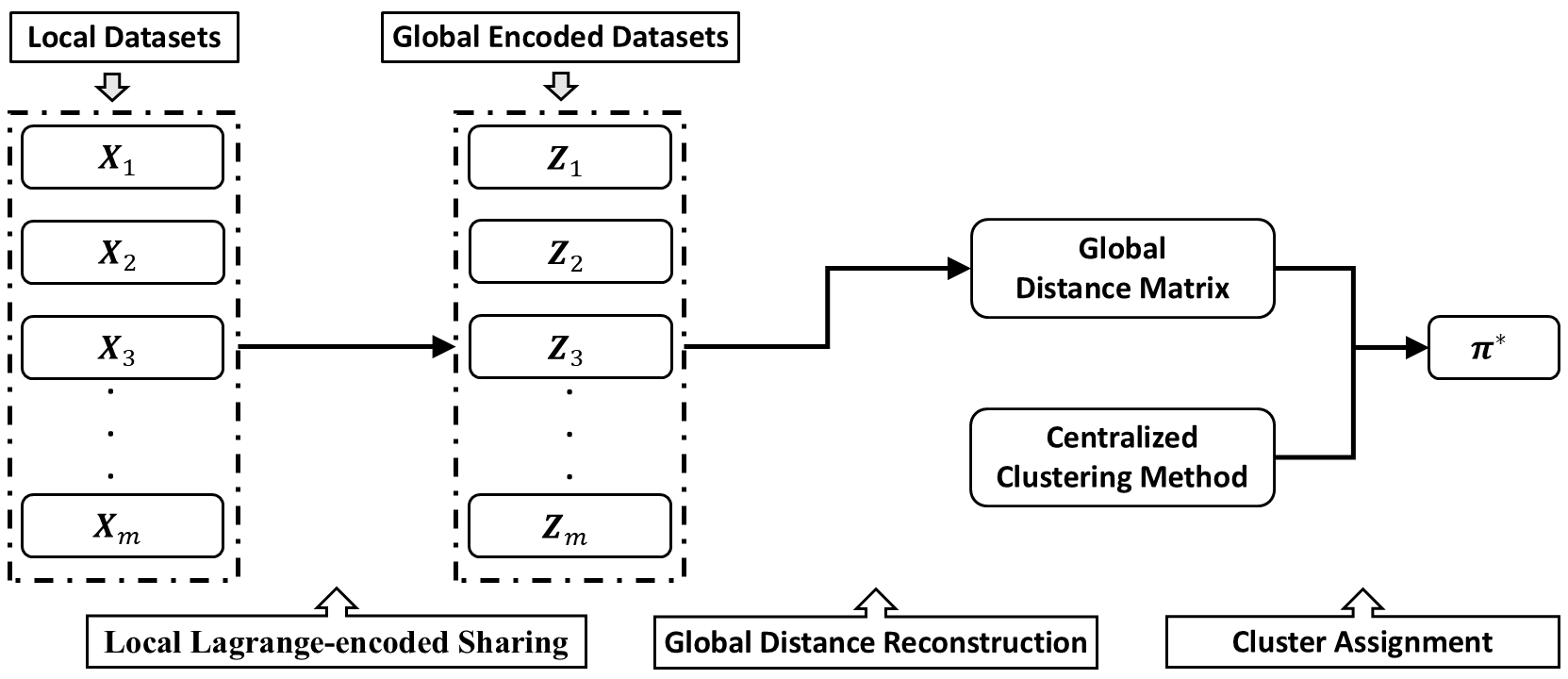}
\caption{\textbf{An overview of the proposed OmniFC.} The architecture comprises three main steps: \textbf{1) Local Lagrange-Encoded Sharing.} Each client $j$ $(j \in [m])$ encodes its private data using Lagrange polynomial interpolation and distributes the encoded data to all peers, enabling each client to construct a global encoded dataset while preserving data privacy. \textbf{2) Global Distance Reconstruction.} Each client $j$ computes pairwise distances within its global encoded dataset and transmits the results to the central server, which leverages them to reconstruct the global distance matrix. \textbf{3) Cluster Assignment.} A centralized clustering method (e.g., k-means) is applied to the global distance matrix to produce the final clustering result $\pi^*$.}
\label{framework}
\end{figure}

\paragraph{Framework Overview.}
As shown in Fig. \ref{framework}, OmniFC comprises three main steps: local Lagrange-encoded sharing, global distance reconstruction, and cluster assignment. Each client $j$ $(j \in [m])$ initially encrypts its local data using Lagrange coded computing (LCC) \cite{yu2019lagrange}, shares the encoded data with peers for pairwise distance computation, and subsequently transmits the resulting distances to the central server for constructing the global distance matrix. Finally, the global distance matrix can serve as input to centralized clustering methods for performing cluster assignment. 


\subsection{OmniFC}

\paragraph{Local Lagrange-encoded Sharing.}
First, each sample $\boldsymbol{x}_i \in \boldsymbol{X}$ $(i \in [n])$—regardless of the client to which it is distributed to—is independently transformed from the real domain $\mathbb{R}^{d}$ to the finite field $\mathbb{F}_p^{d}$ to ensure numerical stability in secure computation \cite{yu2019lagrange}, with $p$ denoting a prime. The transformation is defined as:
\begin{equation}\label{r2f}
\boldsymbol{\tilde{x}}_i = round(2^q\cdot \boldsymbol{x}_i) + p\cdot \frac{|sign(\boldsymbol{x}_i)| - sign(\boldsymbol{x}_i)}{2},
\end{equation}
where $q \in \mathbb{Z}$ regulates the quantization loss. $round(\cdot)$ and $sign(\cdot)$ represent element-wise rounding and sign functions, respectively. Rounding discretizes continuous values to ensure finite field compatibility, while the sign function facilitates correct mapping of negative values \cite{shao2022dres}. We denote the transformed form of $\boldsymbol{X} \in \mathbb{R}^{n\times d}$ as $\boldsymbol{\tilde{X}}\in \mathbb{F}_p^{n\times d}$.

\begin{figure}[!t]
\centering
\includegraphics[height = 6cm, width = 11cm]{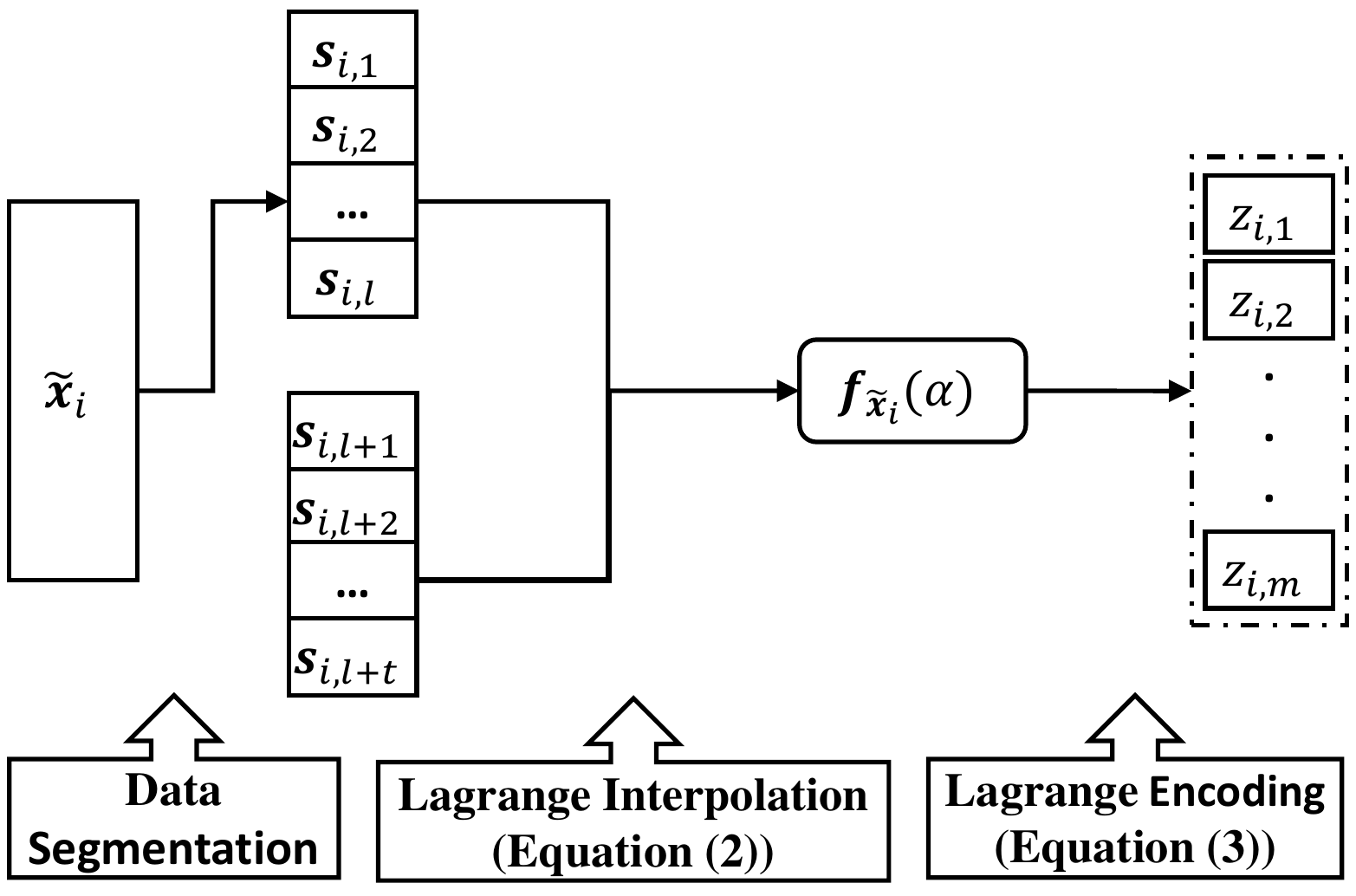}
\caption{\textbf{An illustration of the Lagrange encoding.} Each sample $\boldsymbol{\tilde{x}}_i$ $(i \in [n])$ is initially divided into $l$ segments $\{\boldsymbol{s}_{i, o}\}_{o = 1}^{l}$. Incorporating $t$ additional noises $\{\boldsymbol{s}_{i, l+o}\}_{o = 1}^{t}$, Lagrange interpolation is then conducted as per Equation (\ref{get_f1}) to yield $\boldsymbol{f}_{\boldsymbol{\tilde{x}}_i}(\alpha)$. Subsequently, the encoded representations $\{\boldsymbol{z}_{i, j} \}_{j = 1}^{m}$ of $\boldsymbol{\tilde{x}}_i$ are computed according to Equation (\ref{get_z}).}
\label{lencoding}
\end{figure}

Then, each sample $\boldsymbol{\tilde{x}}_i \in \boldsymbol{\tilde{X}}$ $(i \in [n])$ is independently encoded via  Lagrange polynomial interpolation by the client (Fig. \ref{lencoding}), enabling secret sharing among clients. Specifically, $\boldsymbol{\tilde{x}}_i \in \mathbb{F}_p^{d}$ is partitioned into $l$ segments $\{\boldsymbol{s}_{i,o}\}_{o = 1}^{l}$, i.e., $\boldsymbol{\tilde{x}}_i = [\boldsymbol{s}_{i,1}^T, \boldsymbol{s}_{i,2}^T, \cdots, \boldsymbol{s}_{i,l}^T]^T$, and combined with $t$ random noises to construct a polynomial that serves to encode $\boldsymbol{\tilde{x}}_i$. The noises are introduced to ensure privacy protection against potential client collusion \cite{yu2019lagrange}. Assuming that $d$ is divisible by $l$, the client holds data segments $\boldsymbol{s}_{i,o} \in \mathbb{F}_p^{\frac{d}{l}}$ $(o \in [l])$, and samples $t$ additional noises $\boldsymbol{s}_{i, l+o}$ $(o \in [t])$ uniformly from $\mathbb{F}_p^{\frac{d}{l}}$. Based on the segments $\{\boldsymbol{s}_{i, o}\}_{o = 1}^{l + t}$, the Lagrange interpolation polynomial $\boldsymbol{f}_{\boldsymbol{\tilde{x}}_i}: \mathbb{F}_p \rightarrow \mathbb{F}_p^{\frac{d}{l}} $ of degree $l + t - 1$ can be constructed as follows:
\begin{equation}\label{get_f1}
\boldsymbol{f}_{\boldsymbol{\tilde{x}}_i}(\alpha)=\sum_{o=1}^{l+t} \boldsymbol{s}_{i,o} \cdot \prod_{{o'}\neq o}\frac{\alpha - \alpha_{o'}}{\alpha_o - \alpha_{o'}},
\end{equation}
where $\{{\alpha}_o \}_{o = 1}^{l+t}$ denotes a set of $l + t$ distinct hyperparameters from $\mathbb{F}_p$, pre-specified through agreement among all clients and the central server. Particularly, each data segment $\boldsymbol{s}_{i, o}$ $(o \in [l])$ can be recovered by setting $\alpha = {\alpha}_o$, i.e., $\boldsymbol{f}_{\boldsymbol{\tilde{x}}_i}(\alpha_o) = s_{i, o}$. Beyond the $\{{\alpha}_o \}_{o = 1}^{l+t}$ employed in constructing the polynomial $\boldsymbol{f}_{\boldsymbol{\tilde{x}}_i}$, all clients and the central server also pre-select $m$ distinct public hyperparameters $\{{\beta}_j \}_{j = 1}^{m}$ for encoding, where ${\beta}_j \in \mathbb{F}_p$ and $\{{\alpha}_o \}_{o = 1}^{l+t} \cap \{{\beta}_j \}_{j = 1}^{m} = \emptyset$. Based on $\{{\beta}_j \}_{j = 1}^{m}$, the client encodes its local data $\boldsymbol{\tilde{x}}_i$ into $m$ distinct representations $\{\boldsymbol{z}_{i, j} \}_{j = 1}^{m}$ for secret sharing, with each representation
\begin{equation}\label{get_z}
\boldsymbol{z}_{i,j} = \boldsymbol{f}_{\boldsymbol{\tilde{x}}_i}(\beta_j)
\end{equation} delivered to the $j$-th client.

As these operations are defined per sample, they are universally applicable to local data across all clients. Hence, each client $j$ $(j \in [m])$ will possess a global encoded dataset $\boldsymbol{Z}_j\in \mathbb{F}_p^{n\times \frac{d}{l}}$ corresponding to $\boldsymbol{\tilde{X}}\in \mathbb{F}_p^{n\times d}$, where $\boldsymbol{Z}_j = [\boldsymbol{z}_{1,j}, \boldsymbol{z}_{2,j}, \cdots, \boldsymbol{z}_{n,j}]^T = [\boldsymbol{f}_{\boldsymbol{\tilde{x}}_1}(\beta_j), \boldsymbol{f}_{\boldsymbol{\tilde{x}}_2}(\beta_j), ..., \boldsymbol{f}_{\boldsymbol{\tilde{x}}_n}(\beta_j)]^T$.

\paragraph{Global Distance Reconstruction.}
For each client $j$ $(j \in [m])$, pairwise distances between all encoded representations $\boldsymbol{z}_{i,j}$ and $\boldsymbol{z}_{i',j}$ in $\boldsymbol{Z}_j$ $(i, i' \in [n])$ are calculated and subsequently sent to the central server for constructing the global distance matrix. Specifically, the pairwise distance between $\boldsymbol{z}_{i,j}$ and $\boldsymbol{z}_{i',j}$ can be calculated as:
\begin{equation} \label{get_dis}
{dis}(\boldsymbol{z}_{i,j}, \boldsymbol{z}_{i',j}) = \left\|\boldsymbol{z}_{i,j} - \boldsymbol{z}_{i',j}\right\|_{2}^{2}.
\end{equation}
Based on the $m$ distances $\{{dis}(\boldsymbol{z}_{i,j}, \boldsymbol{z}_{i',j}) \}_{j = 1}^{m}$ provided by the clients, the server can accurately recover the pairwise distance between the corresponding samples \( \boldsymbol{\tilde{x}}_i \) and \( \boldsymbol{\tilde{x}}_{i'} \), as demonstrated in Theorem \ref{the1}.

\begin{theorem}
Let $ \boldsymbol{f}_{\boldsymbol{\tilde{x}}_i, \boldsymbol{\tilde{x}}_{i'}}(\beta): \mathbb{F}_p \rightarrow \mathbb{F}_p $ denote the Lagrange interpolation polynomial interpolated from the set $\{(\beta_j, {dis}(\boldsymbol{z}_{i,j}, \boldsymbol{z}_{i',j}) \}_{j = 1}^{m}$:
\begin{equation}\boldsymbol{f}_{\boldsymbol{\tilde{x}}_i, \boldsymbol{\tilde{x}}_{i'}}(\beta)=\sum_{j=1}^{m} {dis}(\boldsymbol{z}_{i,j}, \boldsymbol{z}_{i',j}) \cdot \prod_{j^{\prime} \neq j} \frac{\beta-\beta_{j^{\prime}}}{\beta_{j}-\beta_{j^{\prime}}},
\end{equation}
where \( \boldsymbol{z}_{i,j} \) and \( \boldsymbol{z}_{i',j} \) denote the encoded representations of arbitrary samples \( \boldsymbol{\tilde{x}}_i \) and \( \boldsymbol{\tilde{x}}_{i'} \) distributed among clients. When $ m \geq 2l + 2t - 1 $, the distance $\text{dis}(\boldsymbol{\tilde{x}}_i, \boldsymbol{\tilde{x}}_{i'})$ can be precisely recovered:
\begin{equation} \label{get_dis_real}
{dis}(\boldsymbol{\tilde{x}}_i, \boldsymbol{\tilde{x}}_{i'}) = \sum_{o=1}^l \boldsymbol{f}_{\boldsymbol{\tilde{x}}_i, \boldsymbol{\tilde{x}}_{i'}}(\alpha_o),
\end{equation}
irrespective of how data is distributed among clients.
\label{the1}
\end{theorem}

\begin{remark}
The condition \( m \geq 2l + 2t - 1 \) imposes minimal practical constraint, given that \( m \) is predefined by the system while \( l \) and \( t \) are tunable hyperparameters. This flexibility allows the condition to be met easily, ensuring the theorem’s practical applicability and highlighting its relevance to real-world implementations.
\end{remark}

Then, by converting \( {dis}(\boldsymbol{\tilde{x}}_i, \boldsymbol{\tilde{x}}_{i'}) \) from the finite field $\mathbb{F}_p$ back to the real domain $\mathbb{R}$, the server recovers:
\begin{equation} \label{f2r}
\left.{dis}(\boldsymbol{x}_i, \boldsymbol{x}_{i'}) =\left\{
\begin{array}
{lll}\frac{1}{2^q} \cdot  {dis}(\boldsymbol{\tilde{x}}_i, \boldsymbol{\tilde{x}}_{i'})
& \mathrm{if} & 0\leq {dis}(\boldsymbol{\tilde{x}}_i,\boldsymbol{\tilde{x}}_{i'})<\frac{p-1}{2} \\[10pt]
\frac{1}{2^q} \cdot  ({dis}(\boldsymbol{\tilde{x}}_i, \boldsymbol{\tilde{x}}_{i'}) - p)
& \mathrm{if} & \frac{p-1}{2}\leq {dis}(\boldsymbol{\tilde{x}}_i,\boldsymbol{\tilde{x}}_{i'})<p
\end{array}\right.\right..
\end{equation}
Based on the recovered distances, we denote the global distance matrix as \( \boldsymbol{D} \in \mathbb{R}^{n \times n} \), with each entry defined as \( \boldsymbol{D}_{ii'} = {dis}(\boldsymbol{x}_i, \boldsymbol{x}_{i'}) \) for \( i, i' \in [n] \).

\paragraph{Cluster Assignment.}
With the recovered global distance matrix \( \boldsymbol{D} \in \mathbb{R}^{n \times n} \), the server can directly perform clustering without requiring any modification to existing centralized clustering methods. This characteristic demonstrates the \textbf{simplicity} and \textbf{flexibility} of the proposed OmniFC framework.

Specifically, pairwise-distance-dependent centralized clustering methods—such as spectral clustering (SC) \cite{ng2001spectral}, DBSCAN \cite{ester1996density}, hierarchical clustering (HC) \cite{gowda1978agglomerative}, and k-medoids (KMed) \cite{rdusseeun1987clustering}—can seamlessly utilize \( \boldsymbol{D} \) for model training, owing to their intrinsic reliance on pairwise sample distances during the clustering process. For methods that do not explicitly depend on pairwise relationships—such as k-means (KM) \cite{macqueen1967some}, fuzzy c-means (FCM) \cite{bezdek1984fcm}, and nonnegative matrix factorization (NMF) \cite{lee2000algorithms}—the server employs \( \boldsymbol{D} \) as a proxy for the raw features \( \boldsymbol{X} \in \mathbb{R}^{n \times d}\) to perform clustering, i.e., the distance values become the new features of the samples. This allows these algorithms to operate as if on centralized data, while implicitly leveraging the global structure encoded in \( \boldsymbol{D} \). These federated extensions of centralized methods built upon OmniFC are denoted as OmniFC-SC, OmniFC-DBSCAN, OmniFC-HC, OmniFC-KMed, OmniFC-KM, OmniFC-FCM, and OmniFC-NMF, respectively. Algorithm \ref{ps_code} in the appendix delineates the pseudocode of OmniFC.

\subsection{Privacy Analysis}
OmniFC adopts LCC encryption to enhance clustering performance while fortifying data privacy. Although LCC enables clients to obtain global awareness via inter-client sharing of Lagrange-encoded data, it also poses emerging privacy threats, as colluding clients may leverage the shared information to infer others’ private data \cite{yu2019lagrange}. Hence, evaluating OmniFC’s resistance to client collusion is essential for delineating its practical applicability. Theorem \ref{the2} provides a formal guarantee that each data point \( \boldsymbol{\tilde{x}}_i \) maintains information-theoretic security in the presence of up to \( t \) colluding clients, thereby affirming the practical applicability of OmniFC.

\begin{theorem}
Given the number of noises $t$, a $t$-private OmniFC is achievable if $ m \geq 2l + 2t - 1$, i.e.,
\begin{equation}
I(\boldsymbol{\tilde{x}}_i;\{\boldsymbol{z}_{i,j}\}_{j \in \boldsymbol{\mathcal{C}}}) = 0,
\end{equation}
where $I(\cdot;\cdot)$ denotes the mutual information function, $\boldsymbol{\mathcal{C}} \subset [m]$ and $|\boldsymbol{\mathcal{C}}| \leq t$.
\label{the2}
\end{theorem}

\begin{remark}
Mutual information essentially measures how much one piece of information reveals about another—when it equals zero, it means one reveals nothing about the other, thereby preserving privacy. Furthermore, mutual information is not an isolated privacy-preserving metric; it can be compared with other privacy measures (such as differential privacy) within a unified framework \cite{wang2016relation}.
\end{remark}

\begin{remark}
The condition for achieving $t$-private security in Theorem \ref{the2} coincides with that for exact distance reconstruction in Theorem \ref{the1}, i.e., $m \geq 2l + 2t - 1$. Consequently, by adhering to this constraint, we can increase the number of noises $t$ to strengthen privacy protection without compromising the precision of distance reconstruction.
\end{remark}

\subsection{Complexity Analysis}

Recall that reconstructing the global pairwise distance matrix involves two main stages: local encoding and distance computation on the client side, followed by server-side decoding to recover the global matrix. For each client $j$, fast polynomial interpolation and evaluation \cite{kedlaya2011fast} yield an encoding complexity of $\mathcal{O}(\frac{dn_jm\log^2m}{l})$, while distance computation on the encoded data incurs a complexity of $\mathcal{O}(\frac{dn^2}{l})$, leading to an overall complexity of $\mathcal{O}(\frac{d(n^2 + n_jm\log^2m)}{l})$. $n_j$ is the number of samples held by client $j$. For the server side, the decoding operation has a complexity of $\mathcal{O}(n^2(l+t)\log^2(l+t))$.

The analysis indicates that, on the client side, increasing the number of data segments $l$ substantially reduces computational complexity, thereby facilitating the calculation of pairwise distances in a $\frac{d}{l}$-dimensional space, where the feature count ($\frac{d}{l}$) involved in pairwise comparisons decreases as $l$ rises. On the server side, however, a larger $l$ results in greater computational complexity. As for the impact of $l$ on the overall complexity of the reconstruction process, a theoretical analysis is challenging due to the differing hyperparameters involved on both sides.

To handle this, we presented the runtime for reconstructing the global pairwise distance matrix across varying data scales on 10x\_73k. As shown in Table \mbox{\ref{tab:rebuttal_time}}, one can observe that: Although the runtime exhibits a marked increase as $n$ grows, enhancing the number of data segments $l$ can considerably boost computational efficiency when $n$ is fixed. For instance, processing 70k samples takes 4352 seconds with $l = 8$, demonstrating that our method remains computationally feasible on large-scale datasets. Importantly, as long as the constraints $m \geq 2(l + t - 1) + 1$ holds, increasing $l$ have no effect on the reconstruction effectiveness of the global pairwise distance matrix (Theorem \ref{the1} and Fig. \ref{dis_sensitivity}).

\section{Experiments}

\subsection{Experimental Setup}
\label{app_details}
\paragraph{Datasets and Evaluation Criteria.}
The proposed OmniFC is assessed using seven benchmark datasets across tabular, visual, temporal, and genomic domains, including Iris \cite{fisher1936use}, MNIST \cite{deng2012mnist}, Fashion-MNIST \cite{xiao2017fashion}, COIL-20 \cite{nene1996columbia}, COIL-100 \cite{nene1996columbia}, Pendigits \cite{keller2012hics}, and 10x\_73k \cite{zheng2017massively}. The chosen datasets encompass diverse modalities, dimensionalities, and cluster patterns, facilitating a comprehensive evaluation of the method's generalizability in practical scenarios.

The evaluation criteria encompass Normalized Mutual Information (NMI) \cite{strehl2002cluster} and Kappa \cite{liu2019evaluation}, with higher scores indicating improved clustering performance. Despite the widespread use of NMI, increasing evidence suggests it may be misleading, whereas Kappa is more reliable \cite{liu2019evaluation, yan2024sda, yan2025significance}. Hence, our analysis is grounded in Kappa-based results, with NMI-based outcomes relegated to the appendix for reference. Details of datasets and evaluation criteria are provided in Appendix \ref{app_de}.

\paragraph{Baselines.}
OmniFC is evaluated in comparison with the federated extensions of several centralized clustering methods, including KM-based (k-FED \cite{dennis2021heterogeneity}, MUFC \cite{pan2023machine}), FCM-based (FFCM \cite{stallmann2022towards}), SC-based (FedSC \cite{qiao2023federated}), and NMF-based (FedMAvg \cite{wang2022federated}) methods. To contextualize the performance of federated clustering against its centralized counterpart, we also present results of vanilla KM, FCM, SC, and NMF under centralized settings, referred to as KM\_central, FCM\_central, SC\_central, and NMF\_central, respectively.

\paragraph{Federated Settings.}
Following Ref. \cite{chung2022federated, yan2024sda}, we simulate  diverse federated settings by partitioning the real-world dataset into $k^\star$ subsets—each representing a client—and adjusting the non-IID level $p$, where $k^\star$ denotes the number of true clusters. Specifically, for each client, a fraction \( p \) of its data is sampled from a single cluster, while the remaining \( 1-p \) portion is drawn uniformly across all clusters. As such, \( p = 0 \) recovers the IID setting, whereas \( p = 1 \) induces a maximally skewed distribution, where each client's data is fully concentrated within a single cluster. Since OmniFC is immune to the Non-IID degree, the Non-IID level $p$ is indicated solely during comparisons with the existing FC baselines and omitted elsewhere.

\begin{figure}[!t]
\centering
\subfigure{
\includegraphics[height = 2.cm, width = 2.75cm]{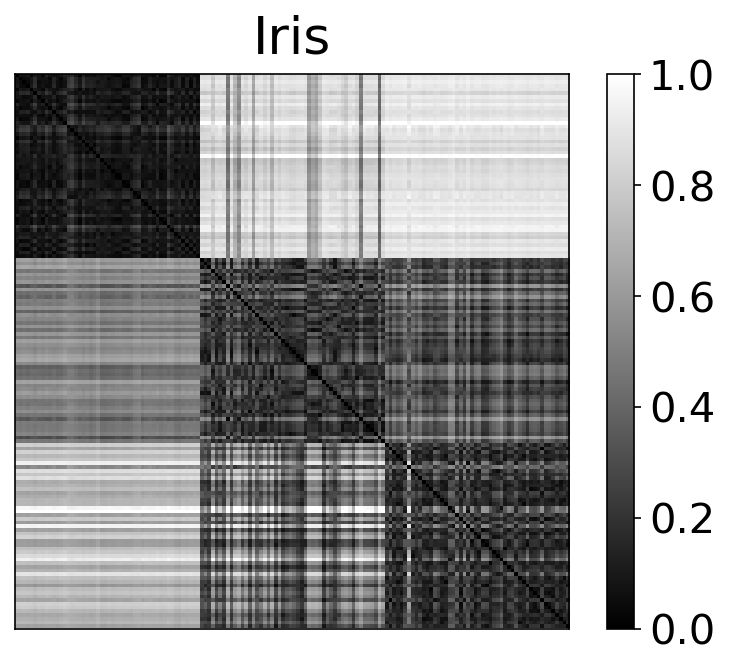}}
\,
\subfigure{
\includegraphics[height = 2.cm, width = 2.75cm]{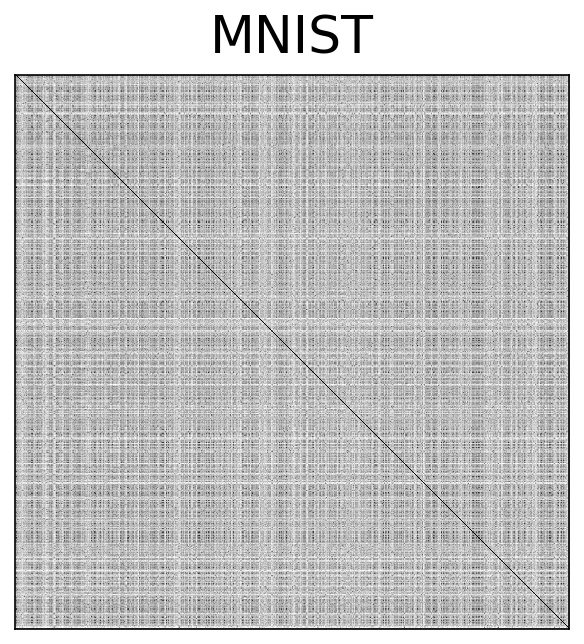}}
\,
\subfigure{
\includegraphics[height = 2cm, width = 2.75cm]{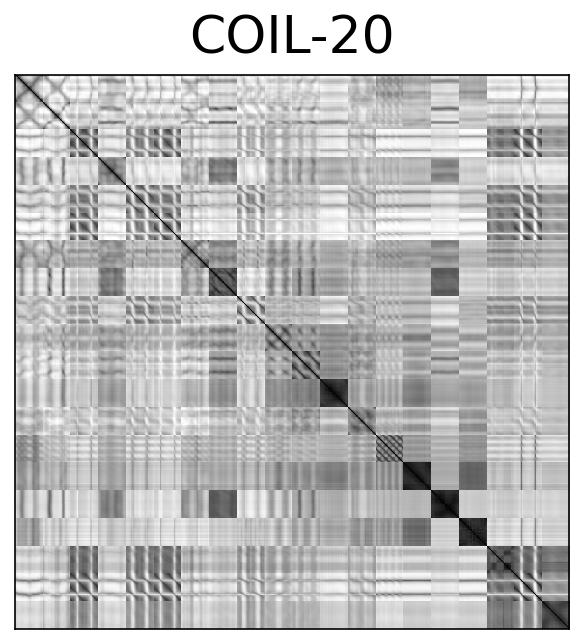}}
\,
\subfigure{
\includegraphics[height = 2cm, width = 2.75cm]{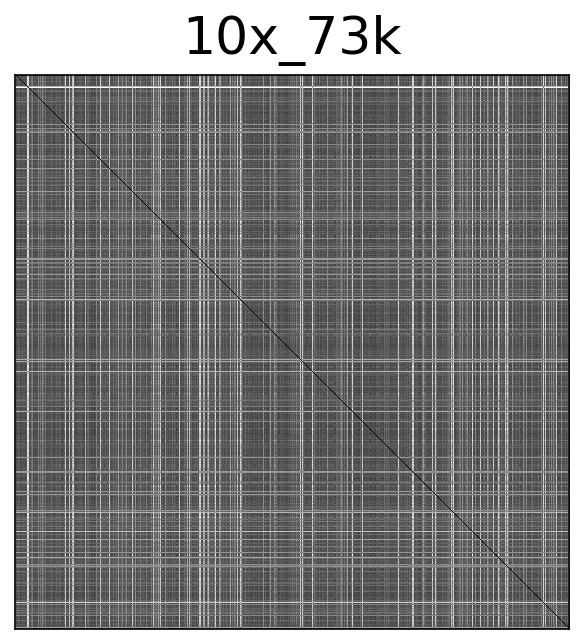}}

\subfigure{
\includegraphics[height = 2.cm, width = 2.75cm]{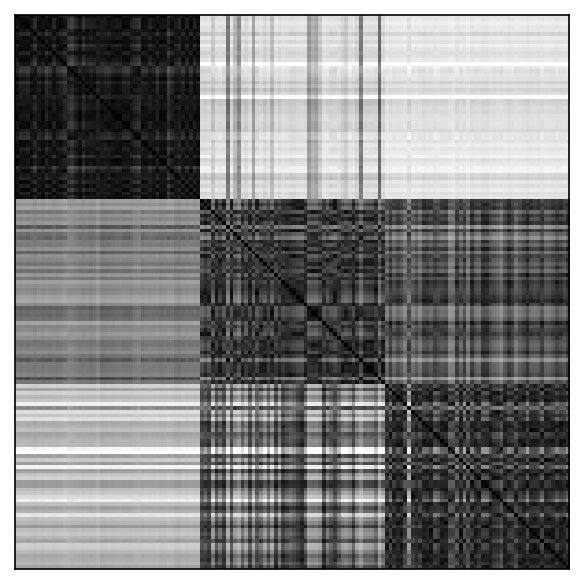}}
\,
\subfigure{
\includegraphics[height = 2.cm, width = 2.75cm]{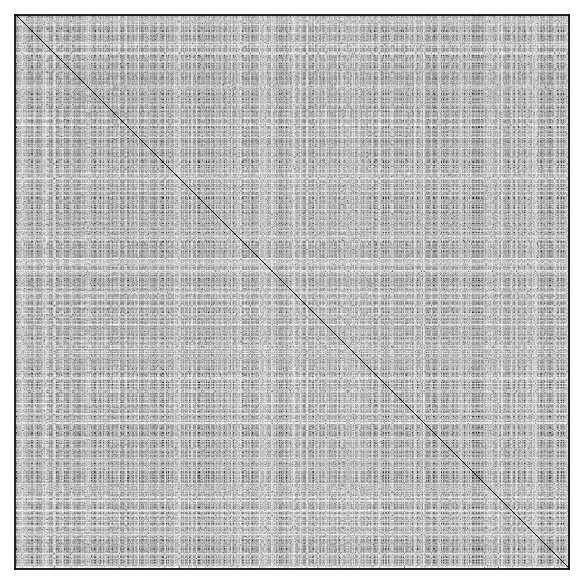}}
\,
\subfigure{
\includegraphics[height = 2.cm, width = 2.75cm]{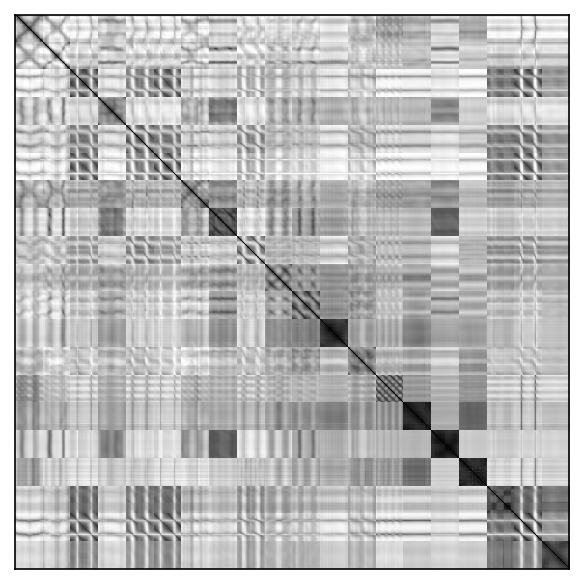}}
\,
\subfigure{
\includegraphics[height = 2.cm, width = 2.75cm]{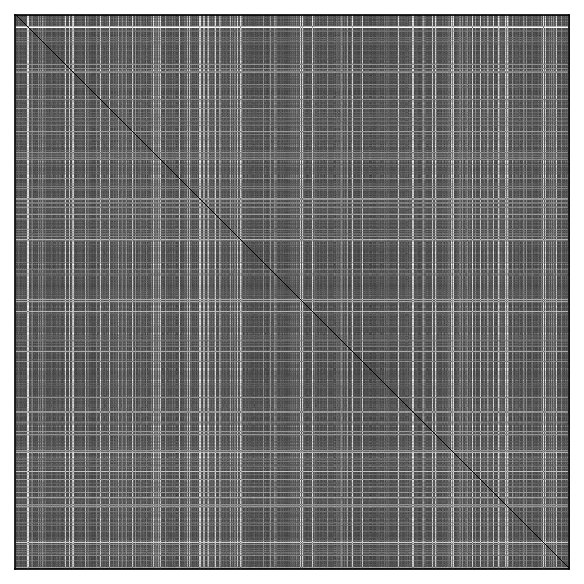}}

\caption{\textbf{Comparison between the ground-truth (top row) and reconstructed (bottom row) pairwise distance matrices.} The visual consistency indicates that the proposed OmniFC faithfully recovers the inter-sample similarity.} 
\label{dis_rec}
\end{figure}

\subsection{Experimental Results}
Our experiments center on three key aspects: 1) the comparative advantage of OmniFC over existing approaches; 2) the generality of OmniFC in extending centralized clustering methods; and 3) the sensitivity of OmniFC to hyperparameters. Implementation details are provided in Appendix \ref{app_id}, and supplementary experimental results are presented in Appendix \ref{app_sr}.

%
%
%

\begin{table}[tbp]
\centering
\caption{\textbf{Kappa of clustering methods in different federated scenarios.} For each comparison, the best result is highlighted in boldface.}
\label{kappa}
\resizebox{\textwidth}{!}{  
\large
\renewcommand{\arraystretch}{1.15} 
\tabcolsep 1mm
\begin{tabular}{llccccccccccccc}
\hline\hline
\multirow{2}{*}{\textbf{Dataset}} &\multirow{2}{*}{$\boldsymbol{p}$} &\multicolumn{3}{c}{\textbf{SC-based methods}} &\multicolumn{4}{c}{\textbf{KM-based methods}} &\multicolumn{3}{c}{\textbf{FCM-based methods}} &\multicolumn{3}{c}{\textbf{NMF-based methods}}\\
\cmidrule(r){3-5} \cmidrule(r){6-9} \cmidrule(r){10-12} \cmidrule(r){13-15}
& & \textcolor{mygray}{\textbf{SC\_central}} & \textbf{FedSC} & \textbf{Ours} & \textcolor{mygray}{\textbf{KM\_central}} & \textbf{k-FED} & \textbf{MUFC} &\textbf{Ours} & \textcolor{mygray}{\textbf{FCM\_central}} & \textbf{FFCM} & \textbf{Ours} & \textcolor{mygray}{\textbf{NMF\_central}} & \textbf{FedMAvg} & \textbf{Ours} \\
\hline
\multirow{5}{*}{Iris}
& 0.00 & \textcolor{mygray}{\multirow{5}{*}{0.95}} & 0.95 & \textbf{0.95} &
\textcolor{mygray}{\multirow{5}{*}{0.95}} & 0.38 & 0.83 & \textbf{0.95} & \textcolor{mygray}{\multirow{5}{*}{0.95}} & \textbf{0.96} & 0.95 & \textcolor{mygray}{\multirow{5}{*}{0.57}} & 0.50 & \textbf{0.95} \\
& 0.25 &  & 0.93 & \textbf{0.95} &  & 0.95 & 0.93 & \textbf{0.95} &  & 0.49 & \textbf{0.95} &  & 0.50 & \textbf{0.95} \\
& 0.50 &  & 0.85 & \textbf{0.95} &  & 0.93 & 0.79 & \textbf{0.95} &  & 0.93 & \textbf{0.95} &  & 0.50 & \textbf{0.95} \\
& 0.75 &  & 0.93 & \textbf{0.95} &  & 0.95 & 0.81 & \textbf{0.95} &  & \textbf{0.96} & 0.95 &  & 0.50 & \textbf{0.95} \\
& 1.00 &  & 0.31 & \textbf{0.95} &  & 0.71 & 0.77 & \textbf{0.95} &  & \textbf{0.97} & 0.95 &  & 0.50 & \textbf{0.95} \\
\midrule
\multirow{5}{*}{MNIST}
& 0.00 & \textcolor{mygray}{\multirow{5}{*}{0.55}} & 0.53 & \textbf{0.55} & \textcolor{mygray}{\multirow{5}{*}{0.47}} & \textbf{0.43} & 0.41 & 0.42 & \textcolor{mygray}{\multirow{5}{*}{0.50}} & \textbf{0.48} & 0.41 & \textcolor{mygray}{\multirow{5}{*}{0.46}} & \textbf{0.40} & 0.38 \\
& 0.25 &  & 0.54 & \textbf{0.55} &  & 0.45 & \textbf{0.50} & 0.42 &  & \textbf{0.52} & 0.41 &  & \textbf{0.44} & 0.38 \\
& 0.50 &  & 0.54 & \textbf{0.55} &  & 0.29 & \textbf{0.46} & 0.42 &  & \textbf{0.53} & 0.41 &  & \textbf{0.39} & 0.38 \\
& 0.75 &  & \textbf{0.58} & 0.55 &  & 0.32 & \textbf{0.47} & 0.42 &  & \textbf{0.45} & 0.41 &  & \textbf{0.45} & 0.38 \\
& 1.00 &  & 0.38 & \textbf{0.55} &  & \textbf{0.47} & 0.43 & 0.42 &  & \textbf{0.48} & 0.41 &  & \textbf{0.46} & 0.38 \\
\midrule
\multirow{5}{*}{Fashion-MNIST}
& 0.00 & \textcolor{mygray}{\multirow{5}{*}{0.53}} & \textbf{0.54} & 0.53 & \textcolor{mygray}{\multirow{5}{*}{0.50}} & 0.46 & 0.43 & \textbf{0.51} & \textcolor{mygray}{\multirow{5}{*}{0.53}} & \textbf{0.51} & 0.50 & \textcolor{mygray}{\multirow{5}{*}{0.51}} & 0.46 & \textbf{0.49} \\
& 0.25 &  & 0.52 & \textbf{0.53} &  & 0.43 & 0.40 & \textbf{0.51} &  & 0.47 & \textbf{0.50} &  & 0.46 & \textbf{0.49} \\
& 0.50 &  & \textbf{0.54} & 0.53 &  & 0.48 & 0.50 & \textbf{0.51} &  & 0.43 & \textbf{0.50} &  & 0.46 & \textbf{0.49} \\
& 0.75 &  & 0.47 & \textbf{0.53} &  & 0.45 & 0.45 & \textbf{0.51} &  & 0.50 & \textbf{0.50} &  & 0.46 & \textbf{0.49} \\
& 1.00 &  & 0.38 & \textbf{0.53} &  & 0.32 & 0.50 & \textbf{0.51} &  & 0.46 & \textbf{0.50} &  & 0.46 & \textbf{0.49} \\
\midrule
\multirow{5}{*}{COIL-20}
& 0.00 & \textcolor{mygray}{\multirow{5}{*}{0.61}} & \textbf{0.68} & 0.63 & \textcolor{mygray}{\multirow{5}{*}{0.64}} & 0.42 & 0.58 & \textbf{0.64} & \textcolor{mygray}{\multirow{5}{*}{0.59}} & 0.51 & \textbf{0.59} & \textcolor{mygray}{\multirow{5}{*}{0.56}} & 0.50 & \textbf{0.61} \\
& 0.25 &  & \textbf{0.68} & 0.63 &  & 0.46 & 0.61 & \textbf{0.64} &  & 0.47 & \textbf{0.59} &  & 0.51 & \textbf{0.61} \\
& 0.50 &  & \textbf{0.73} & 0.63 &  & 0.42 & 0.57 & \textbf{0.64} &  & 0.51 & \textbf{0.59} &  & 0.44 & \textbf{0.61} \\
& 0.75 &  & 0.54 & \textbf{0.63} &  & 0.41 & 0.58 & \textbf{0.64} &  & 0.55 & \textbf{0.59} &  & 0.51 & \textbf{0.61} \\
& 1.00 &  & 0.29 & \textbf{0.63} &  & 0.46 & 0.56 & \textbf{0.64} &  & 0.59 & \textbf{0.59} &  & 0.52 & \textbf{0.61} \\
\midrule
\multirow{5}{*}{COIL-100}
& 0.00 & \textcolor{mygray}{\multirow{5}{*}{0.54}} & 0.34 & \textbf{0.54} & \textcolor{mygray}{\multirow{5}{*}{0.49}} & 0.48 & 0.49 & \textbf{0.56} & \textcolor{mygray}{\multirow{5}{*}{0.49}} & 0.37 & \textbf{0.53} & \textcolor{mygray}{\multirow{5}{*}{0.43}} & 0.39 & \textbf{0.50} \\
& 0.25 &  & 0.32 & \textbf{0.54} &  & 0.45 & 0.50 & \textbf{0.56} &  & 0.38 & \textbf{0.53} &  & 0.38 & \textbf{0.50} \\
& 0.50 &  & 0.32 & \textbf{0.54} &  & 0.41 & 0.48 & \textbf{0.56} &  & 0.49 & \textbf{0.53} &  & 0.39 & \textbf{0.50} \\
& 0.75 &  & 0.29 & \textbf{0.54} &  & 0.41 & 0.50 & \textbf{0.56} &  & 0.48 & \textbf{0.53} &  & 0.39 & \textbf{0.50} \\
& 1.00 &  & 0.27 & \textbf{0.54} &  & 0.43 & 0.52 & \textbf{0.56} &  & 0.51 & \textbf{0.53} &  & 0.38 & \textbf{0.50} \\
\midrule
\multirow{5}{*}{Pendigits}
& 0.00 & \textcolor{mygray}{\multirow{5}{*}{0.72}} & \textbf{0.74} & 0.72 & \textcolor{mygray}{\multirow{5}{*}{0.61}} & 0.59 & 0.59 & \textbf{0.62} & \textcolor{mygray}{\multirow{5}{*}{0.66}} & 0.62 & \textbf{0.66} & \textcolor{mygray}{\multirow{5}{*}{0.45}} & 0.33 & \textbf{0.72} \\
& 0.25 &  & \textbf{0.73} & 0.72 &  & 0.46 & 0.58 & \textbf{0.62} &  & 0.61 & \textbf{0.66} &  & 0.33 & \textbf{0.72} \\
& 0.50 &  & 0.72 & \textbf{0.72} &  & 0.48 & 0.60 & \textbf{0.62} &  & 0.53 & \textbf{0.66} &  & 0.33 & \textbf{0.72} \\
& 0.75 &  & 0.69 & \textbf{0.72} &  & 0.33 & 0.49 & \textbf{0.62} &  & 0.49 & \textbf{0.66} &  & 0.33 & \textbf{0.72} \\
& 1.00 &  & 0.52 & \textbf{0.72} &  & 0.53 & \textbf{0.62} & 0.62 &  & \textbf{0.70} & 0.66 &  & 0.33 & \textbf{0.72} \\
\midrule
\multirow{5}{*}{10x\_73k}
& 0.00 & \textcolor{mygray}{\multirow{5}{*}{0.89}} & 0.52 & \textbf{0.89} & \textcolor{mygray}{\multirow{5}{*}{0.85}} & 0.40 & \textbf{0.63} & 0.56 & \textcolor{mygray}{\multirow{5}{*}{0.53}} & 0.46 & \textbf{0.55} & \textcolor{mygray}{\multirow{5}{*}{0.88}} & 0.49 & \textbf{0.82} \\
& 0.25 &  & 0.52 & \textbf{0.89} &  & 0.55 & \textbf{0.63} & 0.56 &  & 0.47 & \textbf{0.55} &  & 0.49 & \textbf{0.82} \\
& 0.50 &  & 0.52 & \textbf{0.89} &  & 0.57 & \textbf{0.62} & 0.56 &  & \textbf{0.72} & 0.55 &  & 0.49 & \textbf{0.82} \\
& 0.75 &  & 0.54 & \textbf{0.89} &  & 0.37 & \textbf{0.65} & 0.56 &  & \textbf{0.64} & 0.55 &  & 0.50 & \textbf{0.82} \\
& 1.00 &  & 0.24 & \textbf{0.89} &  & 0.30 & \textbf{0.79} & 0.56 &  & \textbf{0.64} & 0.55 &  & 0.50 & \textbf{0.82} \\
 \hline
count &- &- &8 &27   &- &2 &9 &24  &- &13 &22 &- &5 &30\\
\hline\hline
\end{tabular}
}
\end{table}

\paragraph{Efficacy Analysis.}
To comprehensively validate the efficacy of OmniFC, we simulate five scenarios per dataset: IID ($p=0$), mildly non-IID ($p=0.25$), moderately non-IID ($p=0.5$), highly non-IID ($p=0.75$), and fully non-IID ($p=1$). As shown in Table \ref{kappa}, the proposed OmniFC enables superior federated extensions for both pairwise-distance-dependent SC and methods that do not explicitly depend on pairwise relationships, such as KM, FCM, and NMF. For SC, our extended results attain centralized-level clustering fidelity while remaining robust to diverse Non-IID conditions, owing to lossless pairwise distance reconstruction, which remains unaffected by non-IID severity (see Theorem \ref{the1} and Figure \ref{dis_rec}). For centralized methods not explicitly reliant on pairwise relationships, our extended results generally match—and occasionally exceed—their performance under centralized settings, indicating that the global distance matrix \( \boldsymbol{D} \) can serve as an effective surrogate for the raw feature matrix \( \boldsymbol{X} \) to perform clustering.

\paragraph{Generality Analysis.}
To assess OmniFC's generalizability in extending centralized clustering methods, we integrate it with three additional methods (KMed, DBSCAN, and HC) that have been well-studied in centralized contexts but remain underexplored in federated settings. Like SC, all three methods perform clustering based on inter-sample pairwise distances. Hence, by utilizing OmniFC’s lossless distance reconstruction, these three methods can be effortlessly integrated into the OmniFC framework to facilitate lossless federated extensions, as shown in Table 2.


\begin{table}[tbp]\label{dbscan}
\centering
\caption{\textbf{Kappa of different clustering methods.} }
\resizebox{\textwidth}{!}{  
\small
\renewcommand{\arraystretch}{0.9} 
\begin{tabular}{lcccccc}
\hline\hline
\multirow{2}{*}{Dataset} & \multicolumn{2}{c}{KMed-based methods} & \multicolumn{2}{c}{DBSCAN-based methods} & \multicolumn{2}{c}{HC-based methods} \\
\cmidrule(r){2-3} \cmidrule(r){4-5} \cmidrule(r){6-7}
& Central & Ours & Central & Ours & Central & Ours \\
\hline
Iris & 0.94 & 0.94 & 0.50 & 0.50 & 0.94 & 0.95 \\
MNIST & 0.31 & 0.30 & 0.21 & 0.21 & 0.41 & 0.41 \\
Fashion-MNIST & 0.42 & 0.42 & 0.14 & 0.14 & 0.45 & 0.45 \\
COIL-20 & 0.41 & 0.42 & 0.58 & 0.58 & 0.45 & 0.45 \\
COIL-100 & 0.34 & 0.34 & 0.37 & 0.37 & 0.48 & 0.48 \\
Pendigits & 0.44 & 0.45 & 0.48 & 0.48 & 0.57 & 0.57 \\
10x\_73k & 0.30 & 0.30 & 0.01 & 0.01 & 0.28 & 0.28 \\
\hline\hline
\end{tabular}}
\end{table}

\begin{figure}[!t]
\centering
\subfigure{
\includegraphics[height = 5cm, width = 6.5cm]{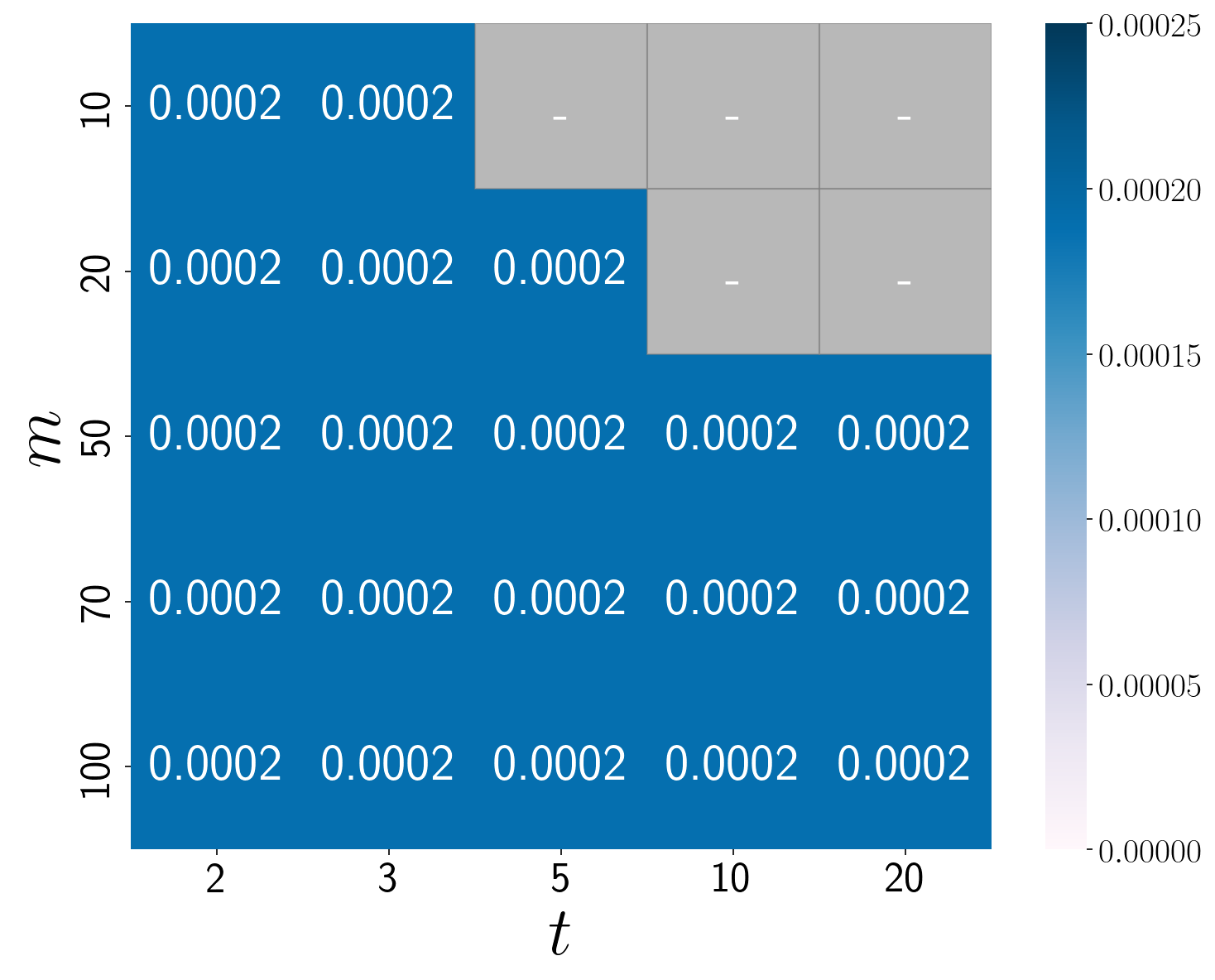}}
\quad
\subfigure{
\includegraphics[height = 5cm, width = 6.5cm]{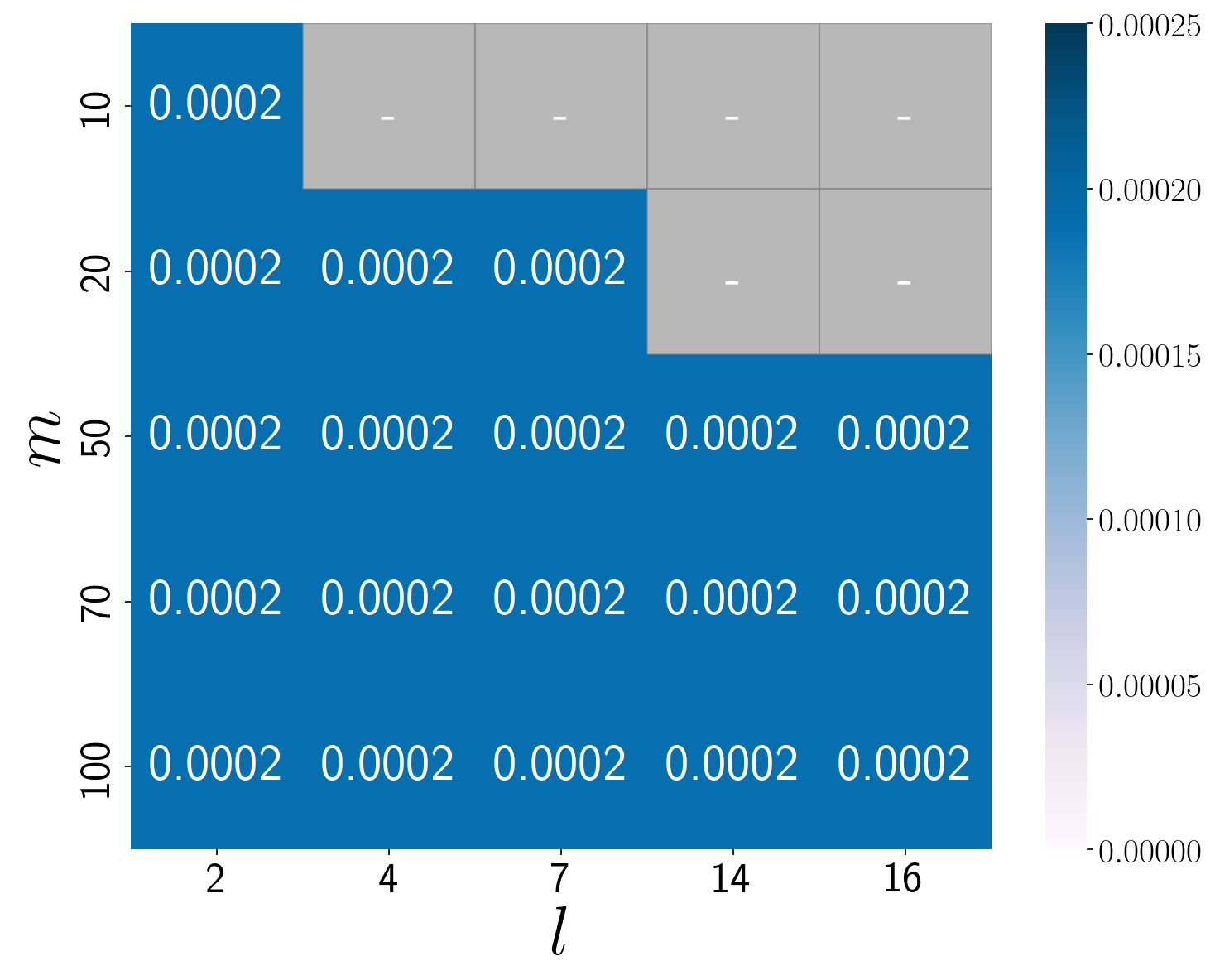}}

\caption{\textbf{Hyperparameter sensitivity of the global distance matrix reconstruction loss.} The gray-highlighted region denotes hyperparameter settings that violate the condition $m \geq 2l + 2t - 1$ in Theorem \ref{the1}, thus precluding distance reconstruction.} 
\label{dis_sensitivity}
\end{figure}

\paragraph{Sensitivity Analysis.}
To assess the hyperparameter sensitivity of OmniFC, we measure the global distance matrix reconstruction loss—defined as the root-mean-square deviation (RMSE) between the ground-truth and reconstructed pairwise distance matrices—across varying number of clients ($m$), noises ($t$), and segments ($l$). In fact, a theoretical guarantee for this has already been provided in Theorem \ref{the1}: as long as the condition \( m \geq 2l + 2t - 1 \) holds, OmniFC is capable of achieving accurate distance reconstruction. This theoretical result is further substantiated by the empirical evidence presented in Fig. \ref{dis_sensitivity}.

\section{Conclusion}

This work introduces OmniFC, a unified and model-agnostic framework via lossless and secure distance reconstruction. Unlike existing methods that rely on model-specific proxies and suffer from data heterogeneity, OmniFC adopts a distance-based perspective that is decoupled from specific clustering models. Benefit from this, theoretical and empirical results show that this framework improves robustness under non-IID settings and supports the extension of a wide range of centralized clustering algorithms to FC.


Beyond FC, the proposed framework may open broader opportunities across federated learning. In particular, the reconstructed global distance matrix can naturally function as a global affinity graph, offering new possibilities for advancing federated graph learning and other domains where capturing global sample relationships is fundamental.

\section*{Acknowledgements}
The authors are grateful to Yong Wang and the three anonymous reviewers for their constructive feedback on an earlier version of this manuscript. Jie Yan is supported by the Strategic Priority Research Program of the Chinese Academy of Sciences, Grant No. XDB1350000.

%

\bibliographystyle{unsrt} 
\bibliography{references}

\newpage
\section*{NeurIPS Paper Checklist}


\begin{enumerate}

\item {\bf Claims}
    \item[] Question: Do the main claims made in the abstract and introduction accurately reflect the paper's contributions and scope?
    \item[] Answer: \answerYes{}
    \item[] Justification: Please refer to the abstract and introduction.

\item {\bf Limitations}
    \item[] Question: Does the paper discuss the limitations of the work performed by the authors?
    \item[] Answer: \answerYes{} 
    \item[] Justification: Please refer to Section \ref{lim}. 

\item {\bf Theory assumptions and proofs}
    \item[] Question: For each theoretical result, does the paper provide the full set of assumptions and a complete (and correct) proof?
    \item[] Answer: \answerYes{} 
    \item[] Justification: Please refer to Section \ref{proof}.

    \item {\bf Experimental result reproducibility}
    \item[] Question: Does the paper fully disclose all the information needed to reproduce the main experimental results of the paper to the extent that it affects the main claims and/or conclusions of the paper (regardless of whether the code and data are provided or not)?
    \item[] Answer: \answerYes{} 
    \item[] Justification: Please refer to Sections \ref{app_details} and \ref{app_id}.
.    

\item {\bf Open access to data and code}
    \item[] Question: Does the paper provide open access to the data and code, with sufficient instructions to faithfully reproduce the main experimental results, as described in supplemental material?
    \item[] Answer: \answerNo{} 
    \item[] Justification: We have provided publicly available dataset information in Section \ref{dataset}. The code for the proposed OmniFC will be released upon the paper's formal publication.

\item {\bf Experimental setting/details}
    \item[] Question: Does the paper specify all the training and test details (e.g., data splits, hyperparameters, how they were chosen, type of optimizer, etc.) necessary to understand the results?
    \item[] Answer: \answerYes{} 
    \item[] Justification: Please refer to Sections \ref{app_details} and \ref{app_id}.

\item {\bf Experiment statistical significance}
    \item[] Question: Does the paper report error bars suitably and correctly defined or other appropriate information about the statistical significance of the experiments?
    \item[] Answer: \answerNo{} 
    \item[] Justification: The deterministic nature of the global distance matrix reconstruction in the proposed OmniFC ensures consistent quality evaluations across multiple runs.

\item {\bf Experiments compute resources}
    \item[] Question: For each experiment, does the paper provide sufficient information on the computer resources (type of compute workers, memory, time of execution) needed to reproduce the experiments?
    \item[] Answer: \answerYes{} 
    \item[] Justification: Please refer to Section \ref{hardware}.

\item {\bf Code of ethics}
    \item[] Question: Does the research conducted in the paper conform, in every respect, with the NeurIPS Code of Ethics \url{https://neurips.cc/public/EthicsGuidelines}?
    \item[] Answer: \answerYes{} 
    \item[] Justification: The research conducted in the paper conform with the NeurIPS Code of Ethics.

\item {\bf Broader impacts}
    \item[] Question: Does the paper discuss both potential positive societal impacts and negative societal impacts of the work performed?
    \item[] Answer: \answerYes{} 
    \item[] Justification: This paper mainly targets privacy-preserving clustering in federated scenarios. By addressing this challenge, we can further promote the practical deployment of clustering in sensitive domains, such as healthcare and finance, while safeguarding data security and user privacy.


\item {\bf Safeguards}
    \item[] Question: Does the paper describe safeguards that have been put in place for responsible release of data or models that have a high risk for misuse (e.g., pretrained language models, image generators, or scraped datasets)?
    \item[] Answer: \answerNA{} 
    \item[] Justification: This paper poses no such risks.

\item {\bf Licenses for existing assets}
    \item[] Question: Are the creators or original owners of assets (e.g., code, data, models), used in the paper, properly credited and are the license and terms of use explicitly mentioned and properly respected?
    \item[] Answer: \answerYes{} 
    \item[] Justification: The original sources of the datasets are cited in Section \ref{app_de}.

\item {\bf New assets}
    \item[] Question: Are new assets introduced in the paper well documented and is the documentation provided alongside the assets?
    \item[] Answer: \answerNA{} 
    \item[] Justification: This paper does not release new assets.

\item {\bf Crowdsourcing and research with human subjects}
    \item[] Question: For crowdsourcing experiments and research with human subjects, does the paper include the full text of instructions given to participants and screenshots, if applicable, as well as details about compensation (if any)?
    \item[] Answer: \answerNA{} 
    \item[] Justification: This paper does not involve crowdsourcing nor research with human subjects.

\item {\bf Institutional review board (IRB) approvals or equivalent for research with human subjects}
    \item[] Question: Does the paper describe potential risks incurred by study participants, whether such risks were disclosed to the subjects, and whether Institutional Review Board (IRB) approvals (or an equivalent approval/review based on the requirements of your country or institution) were obtained?
    \item[] Answer: \answerNA{} 
    \item[] Justification: This paper does not involve crowdsourcing nor research with human subjects.

\item {\bf Declaration of LLM usage}
    \item[] Question: Does the paper describe the usage of LLMs if it is an important, original, or non-standard component of the core methods in this research? Note that if the LLM is used only for writing, editing, or formatting purposes and does not impact the core methodology, scientific rigorousness, or originality of the research, declaration is not required.
    \item[] Answer: \answerNA{} 
    \item[] Justification: The core method development in this research does not involve LLMs as any important, original, or non-standard components.

\end{enumerate}

\newpage
\appendix

\section{Pseudocode of the Proposed OmniFC}
\label{psc}

The procedure of OmniFC is formally presented in Algorithm \ref{ps_code}. On the client side, each sample $\boldsymbol{\tilde{x}}_i$ is independently encoded into $\boldsymbol{z}_{i,j}$ based on Equation (\ref{get_z}), and then transmitted to the $j$-th client, where $i \in [n]$ and $j \in [m]$. Then, each client $j$ computes pairwise distances between all encoded representations $\boldsymbol{z}_{i,j}$ and $\boldsymbol{z}_{i',j}$ $(i, i' \in [n])$ using Equation (\ref{get_dis}), and transmits the results to the central server. On the server side, the global distance matrix is reconstructed based on Equations (\ref{get_dis_real}) and (\ref{f2r}), and subsequently utilized by a centralized clustering algorithm to derive the final clustering outcome $\pi^*$.

\begin{algorithm}[!h]
\caption{OmniFC}\label{ps_code}
\KwIn{Local datasets $\{\boldsymbol{X}_{j} \}_{j = 1}^{m}$, prime number $p$, the number of segments \( l \), the number of noises \( t \), pre-specified hyperparameters $\{{\alpha}_o \}_{o = 1}^{l+t}$ and $\{{\beta}_j \}_{j = 1}^{m}$.}
\KwOut{The final partition $\pi^*$.}
\textbf{Clients execute:}\\
\hspace{1em} \textbf{Local Lagrange Encoding and Secret Sharing:}\\
\hspace{2em} Each sample $\boldsymbol{\tilde{x}}_i$ is encoded via Equation (\ref{get_z}), i.e., $\boldsymbol{z}_{i,j} = \boldsymbol{f}_{\boldsymbol{\tilde{x}}_i}(\beta_j)$, and subsequently \\ \hspace{2.5em} transmitted to the $j$-th client, where $i \in [n]$ and $j \in [m]$. \\
\hspace{1em} \textbf{Global Distance Reconstruction:}\\
\hspace{2em} Each client $j$ computes pairwise distances between all encoded representations $\boldsymbol{z}_{i,j}$ and \\ \hspace{2.5em} $\boldsymbol{z}_{i',j}$ $(i, i' \in [n])$ using Equation (\ref{get_dis}), and transmits the results to the central server. \\
\textbf{Server executes:}\\
\hspace{1em} \textbf{Global Distance Reconstruction:}\\
\hspace{2.3em} The server reconstructs the global distance matrix according to Equations (\ref{get_dis_real}) and (\ref{f2r}).\\

\textbf{Cluster assignment}:\\
\hspace{1em}  The global distance matrix is fed into a centralized clustering method to obtain $\pi^*$.
\end{algorithm}

\section{Proofs of Theorems}
\label{proof}

\begin{table}[!ht]
\center
\caption{\textbf{Notations.}}
\resizebox{\textwidth}{!}{  
\renewcommand{\arraystretch}{2.5} 
\begin{tabular}{ll}
\hline\hline
Notation   &Explanation    \\\hline
$m$
& \parbox{12cm}{
Number of clients.
}\\

$\boldsymbol{X} \in \mathbb{R}^{n\times d}$
& \parbox{12cm}{
The centralized dataset $\boldsymbol{X} \in \mathbb{R}^{n\times d}$ consists of $n$ $d$-dimensional samples $\{\boldsymbol{x}_i\}_{i = 1}^{n}$, which are distributed among $m$ clients, i.e., $\boldsymbol{X}=\bigcup_{j=1}^{m} \boldsymbol{X}_j$.
}\\

$\boldsymbol{\tilde{X}}\in \mathbb{F}_p^{n\times d}$
& \parbox{12cm}{
$\boldsymbol{\tilde{X}}$ denotes the representation of $\boldsymbol{X}$ over the finite field $\mathbb{F}_p^{n\times d}$, consisting of $n$ $d$-dimensional samples $\{\boldsymbol{\tilde{x}}_i\}_{i = 1}^{n}$, where each $\boldsymbol{\tilde{x}}_i$ corresponds to the transformed version of $\boldsymbol{x}_i$ in $\mathbb{F}_p^{d}$.
} \\

$l$
& \parbox{12cm}{
Number of data segments.
}\\

$t$
& \parbox{12cm}{
Number of noises.
}\\

$\boldsymbol{s}_{i,o} \in \mathbb{F}_p^{\frac{d}{l}}$
& \parbox{12cm}{
$\boldsymbol{\tilde{x}}_i = [\boldsymbol{s}_{i,1}^T, \boldsymbol{s}_{i,2}^T, \cdots, \boldsymbol{s}_{i,l}^T]^T$, where $\boldsymbol{s}_{i,o}$ denotes the $o$-th segment of $\boldsymbol{\tilde{x}}_i$ for $o \in [l] = \{1, 2, \cdots, l\}$. For \(l < o \leq l+t\), \(\boldsymbol{s}_{i,o}\) corresponds to the $o$-th noise uniformly sampled from \(\mathbb{F}_p^{\frac{d}{l}}\).
}\\

$\{{\alpha}_o \}_{o = 1}^{l+t}$
& \parbox{12cm}{
A collection of \( l + t \) distinct hyperparameters from \( \mathbb{F}_p \), predetermined by consensus between all clients and the central server, serves to construct the Lagrange interpolation polynomial.
}\\

$\{{\beta}_j \}_{j = 1}^{m}$
& \parbox{12cm}{
A collection of \( m \) distinct hyperparameters from \( \mathbb{F}_p \), predetermined by consensus between all clients and the central server, serves to encode the local data into $m$ distinct representations.
}\\

$\boldsymbol{Z}_j\in \mathbb{F}_p^{n\times \frac{d}{l}}$
& \parbox{12cm}{
The global encoded dataset possessed by client $j$ $(j \in [m]$). $\boldsymbol{Z}_j = [\boldsymbol{z}_{1,j}, \boldsymbol{z}_{2,j}, \cdots, \boldsymbol{z}_{n,j}]^T$, where $\boldsymbol{z}_{i,j}$ $(i \in [n]$) is the encoded representation of $\boldsymbol{\tilde{x}}_i$ at client $j$.
}\\

\hline\hline
\end{tabular}}
\label{notations}
\end{table}

Before proving the theorems, we first summarize some notations used throughout the main text and this appendix, and introduce two lemmas from Ref. \cite{humpherys2020foundations} and \cite{yu2019lagrange}. Refer to Table \ref{notations} for the notions, with the lemmas delineated below. 

\begin{lemma}
\cite{humpherys2020foundations} Given $n$ distinct points $\{(\boldsymbol{x}_{i}, \boldsymbol{y}_{i}) \}_{i = 1}^{n}$ with mutually different $x_i$, there exists a unique polynomial $\boldsymbol{f}(\boldsymbol{x})$ of degree no greater than $n-1$ that interpolates the data, i.e., $\boldsymbol{f}(\boldsymbol{x}_i) = \boldsymbol{y}_i$.
\label{lem1}
\end{lemma}

\begin{lemma}
\cite{yu2019lagrange} Given the number of noises $t$, and a polynomial $\boldsymbol{f}$ used to compute $\boldsymbol{f}(\boldsymbol{\tilde{X}})$, and the degree of $\boldsymbol{f}$ is denoted as $deg(\boldsymbol{f})$. When $ m \geq deg(\boldsymbol{f})(l + t - 1) + 1$, a $t$-private LCC  encryption is achievable, e.g.,
\begin{equation}
I(\boldsymbol{\tilde{x}}_i;\{\boldsymbol{z}_{i,j}\}_{j \in \boldsymbol{\mathcal{C}}}) = 0,
\end{equation}
where $I(\cdot;\cdot)$ denotes the mutual information function, $\boldsymbol{\mathcal{C}} \subset [m]$ and $|\boldsymbol{\mathcal{C}}| \leq t$.
\label{lem2}
\end{lemma}

\paragraph{Proof of Theorem \ref{the1}.}

With Lemma \ref{lem1}, we prove Theorem \ref{the1} as follows.

\begin{proof}
The server possesses only the pre-defined public hyperparameters $\{{\alpha}_o \}_{o = 1}^{l+t}$, $\{{\beta}_j \}_{j = 1}^{m}$ and the distance $\{{dis}(\boldsymbol{z}_{i,j}, \boldsymbol{z}_{i',j} \}_{j = 1}^{m}$. For each distance ${dis}(\boldsymbol{z}_{i,j}, \boldsymbol{z}_{i',j})$ $(j \in [m])$, it can be further formulated as:
\begin{equation}
{dis}(\boldsymbol{z}_{i,j}, \boldsymbol{z}_{i',j}) = \left\|\boldsymbol{z}_{i,j} - \boldsymbol{z}_{i',j}\right\|_{2}^{2} = \left\|\boldsymbol{f}_{\boldsymbol{\tilde{x}}_i}(\beta_j) - \boldsymbol{f}_{\boldsymbol{\tilde{x}}_{i'}}(\beta_j)\right\|_{2}^{2},
\end{equation}
implying that it corresponds to the evaluation of a degree-$2(l + t - 1)$ polynomial at $\beta_j$. According to Lemma \ref{lem1}, the polynomial can be uniquely interpolated from $2(l+t-1) + 1$ distinct points. That is, when \( m \geq 2l + 2t - 1 \), the polynomial can be interpolated from the set \( \{(\beta_j, {dis}(\boldsymbol{z}_{i,j}, \boldsymbol{z}_{i',j})) \}_{j = 1}^{m} \), and it is exactly $\boldsymbol{f}_{\boldsymbol{\tilde{x}}_i, \boldsymbol{\tilde{x}}_{i'}}(\beta)$, i.e.,
\begin{equation}
\boldsymbol{f}_{\boldsymbol{\tilde{x}}_i, \boldsymbol{\tilde{x}}_{i'}}(\beta) = \left\|\boldsymbol{f}_{\boldsymbol{\tilde{x}}_i}(\beta) - \boldsymbol{f}_{\boldsymbol{\tilde{x}}_{i'}}(\beta)\right\|_{2}^{2}.
\label{dis}
\end{equation}
Particularly, by assigning $\beta = \alpha_o$ $(o \in [l])$, the distance between the $o$-th data segments of $\boldsymbol{\tilde{x}}_i$ and $\boldsymbol{\tilde{x}}_{i'}$ can be accurately recovered:
\begin{equation}
\boldsymbol{f}_{\boldsymbol{\tilde{x}}_i, \boldsymbol{\tilde{x}}_{i'}}(\alpha_o) = \left\|\boldsymbol{f}_{\boldsymbol{\tilde{x}}_i}(\alpha_o) - \boldsymbol{f}_{\boldsymbol{\tilde{x}}_{i'}}(\alpha_o)\right\|_{2}^{2} = \left\|\boldsymbol{s}_{i,o} - \boldsymbol{s}_{i',o}\right\|_{2}^{2}.
\end{equation}

Consequently, the distance between $\boldsymbol{\tilde{x}}_i$ and $\boldsymbol{\tilde{x}}_{i'}$ can be precisely reconstructed:
\begin{equation}
\sum_{o=1}^l \boldsymbol{f}_{\boldsymbol{\tilde{x}}_i, \boldsymbol{\tilde{x}}_{i'}}(\alpha_o) = \sum_{o=1}^l \left\|\boldsymbol{s}_{i,o} - \boldsymbol{s}_{i',o}\right\|_{2}^{2} = {dis}(\boldsymbol{\tilde{x}}_i, \boldsymbol{\tilde{x}}_{i'})
\label{dis2}
\end{equation}
Note that since the above proof does not impose any constraints on the distribution of \( \boldsymbol{\tilde{x}}_i \) and \(\boldsymbol{\tilde{x}}_{i'} \) across clients, Equation (\ref{dis2}) holds irrespective of how data is distributed among clients.
\end{proof}

\paragraph{Proof of Theorem \ref{the2}.}
With Lemma \ref{lem2}, we prove Theorem \ref{the2} as follows.
\begin{proof}
We prove Theorem~\ref{the2} by instantiating Lemma~\ref{lem2} with the specific polynomial structure used in the OmniFC framework.

Recall that Lemma~\ref{lem2} states that a $t$-private LCC encryption is achievable when
\[
m \geq \deg(\boldsymbol{f})(l + t - 1) + 1,
\]
where $\boldsymbol{f}$ is the polynomial used in the encoding scheme, and $l$ is the number of data segments.

In the OmniFC setting, the polynomial $\boldsymbol{f}$ is a quadratic distance-based function of degree 2, i.e., $\deg(\boldsymbol{f}) = 2$. Plugging this into the general LCC bound yields:
\[
m \geq 2l + 2t - 1.
\]
Therefore, under this condition, the mutual information between any private input $\boldsymbol{\tilde{x}}_i$ and the encoded messages observed by up to $t$ colluding clients satisfies:
\[
I(\boldsymbol{\tilde{x}}_i;\{\boldsymbol{z}_{i,j}\}_{j \in \boldsymbol{\mathcal{C}}}) = 0,
\]
where $\boldsymbol{\mathcal{C}} \subset [m]$ and $|\boldsymbol{\mathcal{C}}| \leq t$.

This guarantees $t$-privacy in the OmniFC framework, thus completing the proof.
\end{proof}

\begin{table}[!t]
\centering
\caption{\textbf{Reconstruction time (seconds) of the global pairwise distance matrix across diverse data scales under varying $l$.} Computational efficiency can be substantially improved by increasing the number of data segments $l$.}
\renewcommand{\arraystretch}{1.2}
\tabcolsep 5.75mm 
\begin{tabular}{lccccccc}
\hline\hline
$n$ & 1k & 2k & 5k & 10k & 20k & 40k & 70k \\\hline
$l=2$ & 4 & 15 & 78 & 285 & 1053 & 4634 & 13576 \\
$l=4$ & 3 & 9 & 55 & 154 & 620 & 2037 & 7094 \\
$l=8$ & 2 & 6 & 32 & 124 & 435 & 1471 & 4352 \\
\hline\hline
\label{tab:rebuttal_time}
\end{tabular}
\end{table}

\section{Experimental Details}
\label{hardware}

All experiments are implemented in Python and executed on a system equipped with an Intel Core i7-12650H CPU, 16GB of RAM, and an NVIDIA GeForce RTX 4060 GPU.

\subsection{Datasets and Evaluation Criteria}
\label{app_de}

\begin{table}[!t]
\centering
\caption{\textbf{Description of datasets.}}
\resizebox{\textwidth}{!}{  
\renewcommand{\arraystretch}{1.} 
\begin{tabular}{lllll}
\hline\hline
Dataset & Type &Size &Image size/Features & Class \\\hline
Iris            & tabular      & 150      & 4              & 3\\
MNIST           & image        & 70000    & $28\times28$   & 10\\
Fashion-MNIST   & image        & 70000    & $28\times28$   & 10\\
COIL-20         & image        & 1440     & $128\times128$ & 20 \\
COIL-100        & image        & 7200     & $128\times128$ & 100 \\
Pendigits       & time series   & 10992    & 16            & 10\\
10x\_73k        & gene        & 73233    & 720   & 8 \\
\hline\hline
\label{datasets}
\end{tabular}}
\end{table}

\paragraph{Datasets.}
\label{dataset}
As shown in Table \ref{datasets}, we select seven benchmark datasets across tabular, visual, temporal, and genomic domains, including Iris \cite{fisher1936use}, MNIST \cite{deng2012mnist}, Fashion-MNIST \cite{xiao2017fashion}, COIL-20 \cite{nene1996columbia}, COIL-100 \cite{nene1996columbia}, Pendigits \cite{keller2012hics}, and 10x\_73k \cite{zheng2017massively}. The chosen datasets encompass diverse modalities, dimensionalities, and cluster patterns, facilitating a comprehensive evaluation of the method's generalizability in practical scenarios.

\begin{figure}[!t]
\centering
\includegraphics[height = 7cm, width = 14cm]{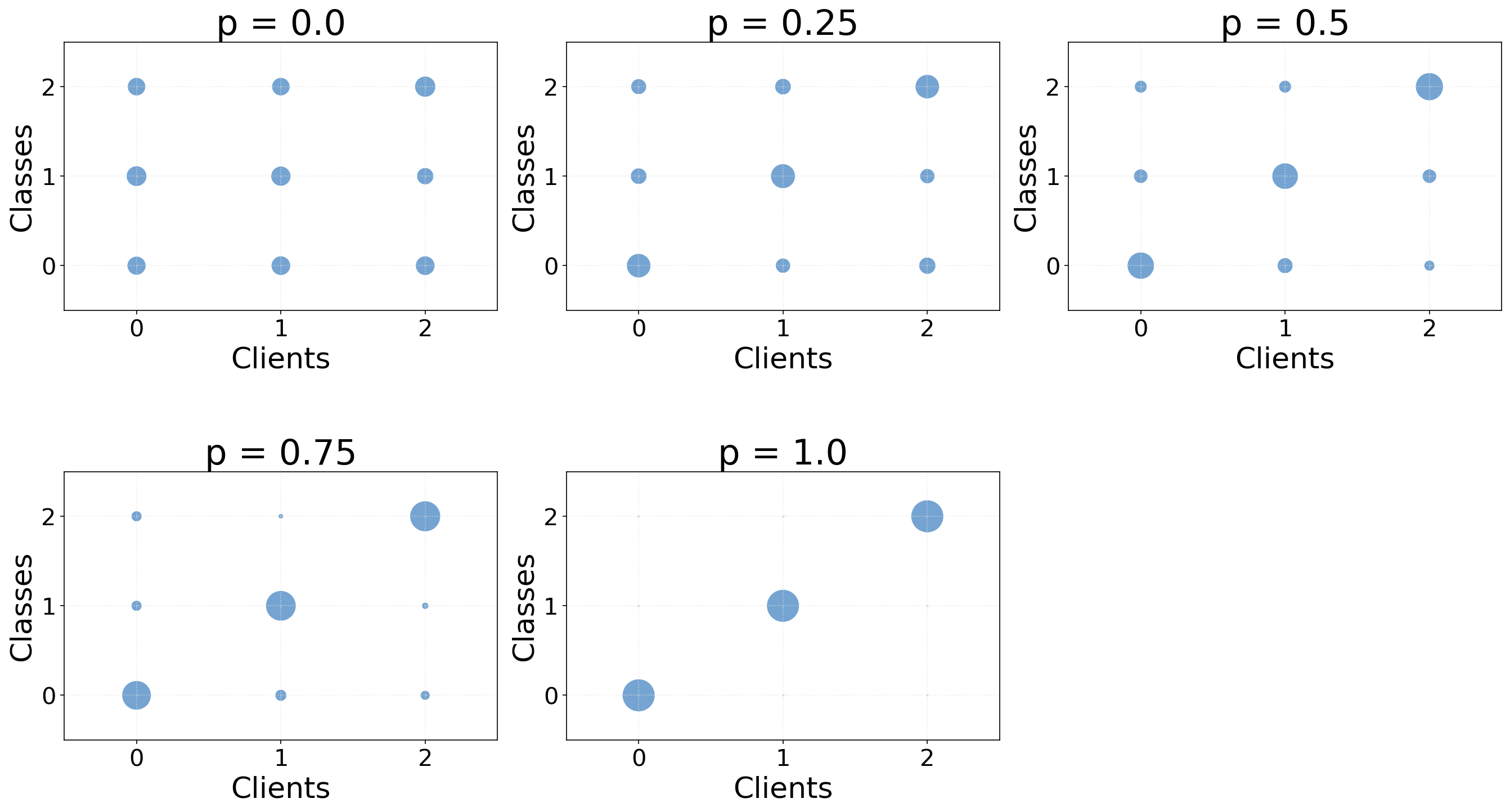}
\caption{\textbf{Data partition visualization on Iris.}}
\label{noniid_vis}
\end{figure}

Fig. \ref{noniid_vis} exemplifies, through the Iris dataset, our simulation of federated scenarios under different Non-IID conditions. We simulate  diverse federated settings by evenly partitioning the Iris dataset into 3 (the number of true clusters) subsets—each representing a client—and adjusting the non-IID level $p$. For a client with $50$ datapoints, the first $p\cdot 50$ datapoints are sampled from a single cluster, and the remaining $(1 - p)\cdot 50$ ones are randomly sampled from any cluster. As such, \( p = 0 \) recovers the IID setting, whereas \( p = 1 \) induces a maximally skewed distribution, where each client's data is fully concentrated within a single cluster.

\paragraph{Evaluation Criteria.}

Evaluation is based on two metrics—normalized mutual information (NMI) \cite{strehl2002cluster} and Kappa \cite{liu2019evaluation}—where elevated scores denote superior clustering quality. Despite being widely adopted, NMI has been shown to have limitations, such as the finite size effect, and fails to account for the importance of small clusters \cite{liu2019evaluation, yan2024sda, yan2025significance}. In contrast, Kappa addresses these concerns, making it a more reliable alternative for clustering evaluation. Hence, our analysis is grounded in Kappa-based results, with NMI-based outcomes serving only as supplementary references.

\subsection{Implementation Details}
\label{app_id}
All centralized clustering methods are implemented by leveraging existing open-source Python libraries: KM, KMed, SC, NMF, and DBSCAN utilize the sklearn library \cite{scikit-learn}, HC employs the scipy library \cite{2020SciPy-NMeth}, and FCM adopts an individual open-source implementation \cite{dias2019fuzzy}. For OmniFC, $\{{\alpha}_o \}_{o = 1}^{l+t}$ is set as a sequence of $l + t$ consecutive odd integers starting from 1, while $\{{\beta}_j \}_{j = 1}^{m}$ is set as a sequence of $m$ consecutive even integers starting from 0. The default values of $l$ and $t$ are set to 2.

We acknowledge the importance of the constraint $m \geq 2(l + t - 1) + 1$ for successful distance reconstruction. Among these hyperparameters, $m$ (number of clients) is typically predetermined by the federated scenario, while $l$ (segments) and $t$ (noises) are tunable hyperparameters that can be adjusted to satisfy this constraint. In practice, violating this condition is extremely rare because the number of clients $m$ in typical federated learning scenarios ranges from tens to millions \cite{li2020federated}, providing substantial flexibility for hyperparameter selection. In practice, $m$ is fixed by the system, while the choice of $l$ and $t$ depends on the relative emphasis placed on communication and computational efficiency versus privacy preservation. Note that as long as $m \geq 2(l + t - 1) + 1$ holds, the specific values of $l$ and $t$ have no effect on the reconstruction effectiveness of the global pairwise distance matrix  (Theorem \ref{the1} and Fig. \ref{dis_sensitivity}). The hyperparameter $l$ governs the dimensionality ($\frac{d}{l}$) of both the shared encoded data across clients and the features involved in pairwise distance computation. A larger $l$ reduces the volume of shared data and the number of features compared, thus improving communication and computational efficiency (Table \ref{tab:rebuttal_time}). The hyperparameter $t$ governs the information-theoretic security, with higher values enhancing resilience against client collusion (Theorem \ref{the2}). However, the constraint $m \geq 2(l + t - 1) + 1$ precludes simultaneous increases in $l$ and $t$, indicating that their selection hinges on the trade-off between communication/computational efficiency and privacy preservation.

Additionally, several clustering methods evaluated in our experiments demand full \( n \times n \) pairwise distance matrix computations, imposing substantial computational and memory burdens on large-scale datasets. To facilitate the execution of comprehensive experiments, we implement a subsampling strategy whereby 1000 samples are randomly drawn from datasets exceeding 5000 entries to form the experimental subset. This approach balances computational efficiency with the preservation of the original data distribution, enabling fair and meaningful comparisons across methods. The sensitivity of the proposed OmniFC with respect to the number of samples is presented in Appendix \ref{app_sca}.

\begin{table}[htbp]
\centering
\caption{\textbf{NMI of clustering methods in different federated scenarios.} For each comparison, the best result is highlighted in boldface.}
\label{nmi}
\resizebox{\textwidth}{!}{  
\large
\renewcommand{\arraystretch}{1.1} 
\tabcolsep 1mm
\begin{tabular}{llccccccccccccc}
\hline\hline
\multirow{2}{*}{\textbf{Dataset}} &\multirow{2}{*}{$\boldsymbol{p}$} &\multicolumn{3}{c}{\textbf{SC-based methods}} &\multicolumn{4}{c}{\textbf{KM-based methods}} &\multicolumn{3}{c}{\textbf{FCM-based methods}} &\multicolumn{3}{c}{\textbf{NMF-based methods}}\\
\cmidrule(r){3-5} \cmidrule(r){6-9} \cmidrule(r){10-12} \cmidrule(r){13-15}
& & \textcolor{mygray}{\textbf{SC\_central}} & \textbf{FedSC} & \textbf{Ours} & \textcolor{mygray}{\textbf{KM\_central}} & \textbf{k-FED} & \textbf{MUFC} &\textbf{Ours} & \textcolor{mygray}{\textbf{FCM\_central}} & \textbf{FFCM} & \textbf{Ours} & \textcolor{mygray}{\textbf{NMF\_central}} & \textbf{FedMAvg} & \textbf{Ours} \\
\hline
\multirow{5}{*}{Iris} & 0.00 & \textcolor{mygray}{\multirow{5}{*}{0.90}} & 0.90 & \textbf{0.90} & \textcolor{mygray}{\multirow{5}{*}{0.90}} & 0.66 & 0.76 & \textbf{0.90} & \textcolor{mygray}{\multirow{5}{*}{0.90}} & \textbf{0.91} & 0.90 & \textcolor{mygray}{\multirow{5}{*}{0.56}} & 0.73 & \textbf{0.90} \\
 & 0.25 &  & 0.85 &\textbf{0.90}  &  & 0.90 & 0.85 &\textbf{0.90}  &  & 0.72 &\textbf{0.90}  &  & 0.73 &\textbf{0.90}  \\
 & 0.50 &  & 0.75  &\textbf{0.90}  &  & 0.87 & 0.70 &\textbf{0.90}  &  & 0.87 &\textbf{0.90}  &  & 0.73 &\textbf{0.90}  \\
 & 0.75 &  & 0.85 &\textbf{0.90}  &  &0.90 & 0.74 &\textbf{0.90}  &  & \textbf{0.91} &0.90  &  & 0.73 &\textbf{0.90}  \\
 & 1.00 &  & 0.29    &\textbf{0.90}  &  & 0.70 & 0.70 &\textbf{0.90}  &  & \textbf{0.93} &0.90  &  & 0.73 &\textbf{0.90}  \\
\hline
\multirow{5}{*}{MNIST} & 0.00
& \textcolor{mygray}{\multirow{5}{*}{0.58}} & \textbf{0.59} & 0.58 & \textcolor{mygray}{\multirow{5}{*}{0.54}} & \textbf{0.51} & 0.48 & 0.46 & \textcolor{mygray}{\multirow{5}{*}{0.55}} & \textbf{0.53} &0.43 &
\textcolor{mygray}{\multirow{5}{*}{0.47}} & \textbf{0.48} & 0.47 \\
 & 0.25 &  & \textbf{0.60} &0.58  &  & 0.49 & \textbf{0.52} &0.46  &  & \textbf{0.53} &0.43  &  & 0.45 &\textbf{0.47}  \\
 & 0.50 &  & \textbf{0.59}  &0.58  &  & 0.39 & \textbf{0.50} &0.46  &  & \textbf{0.52} &0.43  &  & 0.43 &\textbf{0.47}  \\
 & 0.75 &  & \textbf{0.59} &0.58  &  & 0.46 & \textbf{0.52} &0.46  &  & \textbf{0.52} &0.43  &  & 0.47 &\textbf{0.47}  \\
 & 1.00 &  & 0.45    &\textbf{0.58}  &  & 0.51 & \textbf{0.55} &0.46  &  & \textbf{0.57} &0.43  &  & 0.47 &\textbf{0.47}  \\
\hline
\multirow{5}{*}{Fashion-MNIST} & 0.00 & \textcolor{mygray}{\multirow{5}{*}{0.61}} & \textbf{0.61} & 0.61 &
\textcolor{mygray}{\multirow{5}{*}{0.62}} & 0.56 & \textbf{0.56} & 0.52 & \textcolor{mygray}{\multirow{5}{*}{0.61}} & \textbf{0.61} & 0.53 & \textcolor{mygray}{\multirow{5}{*}{0.60}} & 0.53 & \textbf{0.55} \\
 & 0.25 &  & 0.60 &\textbf{0.61}  &  & 0.54 & \textbf{0.54} &0.52  &  & \textbf{0.59} &0.53  &  & 0.53 &\textbf{0.55}  \\
 & 0.50 &  & \textbf{0.61}  &0.61  &  & 0.57 & \textbf{0.60} &0.52  &  & \textbf{0.58} &0.53  &  & 0.53 &\textbf{0.55}  \\
 & 0.75 &  & 0.55 &\textbf{0.61}  &  & \textbf{0.55} & 0.54 &0.52  &  & \textbf{0.61} &0.53 &  & 0.53  &\textbf{0.55}  \\
 & 1.00 &  & 0.39    &\textbf{0.61}  &  & 0.48 & \textbf{0.59} &0.52  &  & \textbf{0.58} &0.53  &  & 0.53 &\textbf{0.55}  \\
\hline
\multirow{5}{*}{COIL-20} & 0.00 & \textcolor{mygray}{\multirow{5}{*}{0.75}} & \textbf{0.80} &0.75 & \textcolor{mygray}{\multirow{5}{*}{0.74}} & 0.65 & \textbf{0.74} & 0.74 & \textcolor{mygray}{\multirow{5}{*}{0.75}} & 0.71 & \textbf{0.72} & \textcolor{mygray}{\multirow{5}{*}{0.70}} & 0.62 & \textbf{0.75} \\
 & 0.25 &  & \textbf{0.78} &0.75  &  & 0.70 & 0.73 &\textbf{0.74}  &  & 0.69 &\textbf{0.72}  &  & 0.62 &\textbf{0.75}  \\
 & 0.50 &  & \textbf{0.80}  &0.75  &  & 0.66 & 0.72 &\textbf{0.74}  &  & \textbf{0.72} &0.72  &  & 0.62 &\textbf{0.75}  \\
 & 0.75 &  & 0.69 &\textbf{0.75}  &  & 0.67 & 0.73 &\textbf{0.74}  &  & \textbf{0.74} &0.72  &  & 0.63 &\textbf{0.75}  \\
 & 1.00 &  & 0.46    &\textbf{0.75}  &  & 0.69 & 0.72 &\textbf{0.74}  &  & \textbf{0.75} &0.72  &  & 0.63 &\textbf{0.75}  \\
\hline
\multirow{5}{*}{COIL-100} & 0.00 & \textcolor{mygray}{\multirow{5}{*}{0.79}} & 0.67 & \textbf{0.79} & \textcolor{mygray}{\multirow{5}{*}{0.77}} & 0.76 & 0.76 & \textbf{0.79} & \textcolor{mygray}{\multirow{5}{*}{0.79}} & 0.69 & \textbf{0.79} & \textcolor{mygray}{\multirow{5}{*}{0.72}} & 0.70 &\textbf{0.76} \\
 & 0.25 &  & 0.66 &\textbf{0.79}  &  & 0.75 & 0.76 &\textbf{0.79}  &  & 0.71 &\textbf{0.79}  &  & 0.70 &\textbf{0.76}  \\
 & 0.50 &  & 0.66  &\textbf{0.79}  &  & 0.75 & 0.76 &\textbf{0.79}  &  & 0.77 &\textbf{0.79}  &  & 0.70 &\textbf{0.76}  \\
 & 0.75 &  & 0.64 &\textbf{0.79}  &  & 0.75 & 0.76 &\textbf{0.79}  &  & 0.77 &\textbf{0.79}  &  & 0.70 &\textbf{0.76}  \\
 & 1.00 &  & 0.61    &\textbf{0.79}  &  & 0.75 &0.79 &\textbf{0.79}  &  & \textbf{0.81} &0.79  &  & 0.70 &\textbf{0.76}  \\
\hline
\multirow{5}{*}{Pendigits} & 0.00 & \textcolor{mygray}{\multirow{5}{*}{0.72}} & \textbf{0.77} & 0.72 & \textcolor{mygray}{\multirow{5}{*}{0.69}} & 0.67 & 0.67 & \textbf{0.67} & \textcolor{mygray}{\multirow{5}{*}{0.69}} & 0.68 & \textbf{0.70} & \textcolor{mygray}{\multirow{5}{*}{0.55}} & 0.42 &\textbf{0.71} \\
 & 0.25 &  & \textbf{0.76} &0.72  &  & 0.62 & 0.66 &\textbf{0.67}  &  & 0.68 &\textbf{0.70}  &  & 0.42 &\textbf{0.71}  \\
 & 0.50 &  & \textbf{0.74}  &0.72  &  & 0.63 & 0.67 &\textbf{0.67}  &  & 0.67 &\textbf{0.70}  &  & 0.42 &\textbf{0.71}  \\
 & 0.75 &  & \textbf{0.75} &0.72  &  & 0.50 & 0.64 &\textbf{0.67}  &  & 0.65 &\textbf{0.70}  &  & 0.42 &\textbf{0.71}  \\
 & 1.00 &  & 0.62    &\textbf{0.72}  &  & 0.64 & \textbf{0.71} &0.67  &  & 0.69 &\textbf{0.70}  &  & 0.42 &\textbf{0.71}  \\
\hline

\multirow{5}{*}{10X\_73k} & 0.00 & \textcolor{mygray}{\multirow{5}{*}{0.85}} & 0.71 & \textbf{0.85} & \textcolor{mygray}{\multirow{5}{*}{0.82}} & \textbf{0.68} & 0.65 & 0.58 & \textcolor{mygray}{\multirow{5}{*}{0.68}} & \textbf{0.69} & 0.58 & \textcolor{mygray}{\multirow{5}{*}{0.83}} & 0.66 &\textbf{0.78} \\
 & 0.25 &  & 0.71 &\textbf{0.85}  &  & \textbf{0.70} & 0.68 &0.58  &  & \textbf{0.70} &0.58  &  & 0.66 &\textbf{0.78}  \\
 & 0.50 &  & 0.70  &\textbf{0.85}  &  & \textbf{0.73} & 0.72 &0.58  &  & \textbf{0.79} &0.58  &  & 0.66 &\textbf{0.78}  \\
 & 0.75 &  & 0.59 &\textbf{0.85}  &  & 0.65 & \textbf{0.73} &0.58  &  & \textbf{0.83} &0.58  &  & 0.66 &\textbf{0.78}  \\
 & 1.00 &  & 0.19    &\textbf{0.85}  &  & 0.49 & \textbf{0.80} &0.58  &  & \textbf{0.82} &0.58  &  & 0.66 &\textbf{0.78}  \\
 \hline
count &- &- &13 &22   &- &5 &12 &18  &- &22 &13 &- &1 &34\\
\hline\hline
\end{tabular}}
\end{table}

\section{Supplementary Experimental Results}
\label{app_sr}

\subsection{NMI-based Evaluation Results}
To supplement the Kappa-based evaluation results and to enable broader comparability with existing FC works, we additionally provide NMI-based evaluation results in Tables \ref{nmi} and \ref{nmi_gen}. Similar to the Kappa-based evaluation results, the numerical results based on NMI also confirm the effectiveness and generalizability of OmniFC.

\subsection{Sensitivity Analysis}
\label{app_sca}

To assess the sensitivity of the proposed OmniFC concerning the number of samples, we evaluate the global distance matrix reconstruction loss—defined as the root-mean-square deviation (RMSE) between the ground-truth and the reconstructed pairwise distance matrices—across different sample sizes. As shown in Table \ref{rmse_n}, OmniFC exhibits favorable scalability concerning sample size.

\begin{table}[tbp]
\centering
\caption{\textbf{NMI of different clustering methods.} }
\label{nmi_gen}
\resizebox{\textwidth}{!}{  
\begin{tabular}{lcccccc}
\hline\hline
\multirow{2}{*}{Dataset} & \multicolumn{2}{c}{KMed-based methods} & \multicolumn{2}{c}{DBSCAN-based methods} & \multicolumn{2}{c}{HC-based methods} \\
\cmidrule(r){2-3} \cmidrule(r){4-5} \cmidrule(r){6-7}
& Central & Ours & Central & Ours & Central & Ours \\
\hline
Iris & 0.86 & 0.86 & 0.73 & 0.73 & 0.89 & 0.90 \\
MNIST & 0.38 & 0.38 & 0.56 & 0.56 & 0.49 & 0.49 \\
Fashion-MNIST & 0.49 & 0.49 & 0.53 & 0.53 & 0.54 & 0.54 \\
COIL-20 & 0.60 & 0.61 & 0.86 & 0.86 & 0.70 & 0.69 \\
COIL-100 & 0.69 & 0.69 & 0.85 & 0.85 & 0.78 & 0.78 \\
Pendigits & 0.56 & 0.55 & 0.74 & 0.74 & 0.69 & 0.69 \\
10x\_73k & 0.31 & 0.31 & 0.45 & 0.45 & 0.51 & 0.50 \\
\hline\hline
\end{tabular}}
\end{table}

\begin{table}[tbp]
\centering
\caption{\textbf{Sample-size sensitivity of the global distance matrix reconstruction loss on MNIST.} }
\label{rmse_n}
\resizebox{\textwidth}{!}{  
\small
\begin{tabular}{lllllll}
\hline\hline
$n$ & 1000 & 2000 & 3000 & 5000 & 7000 & 10000 \\
\hline
RMSE & 0.0002 & 0.0002 & 0.0002 & 0.0002 & 0.0002 & 0.0002 \\
\hline\hline
\end{tabular}}
\end{table}

\subsection{Non-IID Partitioning based on Dirichlet Distributions}

To test more diverse Non-IID scenarios, we used Dirichlet distributions with varying concentrations $\alpha$ to model label heterogeneity on Pendigits in Table \ref{kappa_fed_alpha}, and other experimental settings are consistent with those in Table \ref{kappa}. Experimental results indicate that the OmniFC method exhibits superior efficacy and robustness.

\begin{table}[!ht]
\centering
\caption{\textbf{Kappa of clustering methods in different federated scenarios.} For each comparison, the best result is highlighted in boldface.}
\label{kappa_fed_alpha}
\resizebox{\textwidth}{!}{
\large
\renewcommand{\arraystretch}{2}
\tabcolsep 1mm
\begin{tabular}{lccccccccccccc}
\hline\hline
\multirow{2}{*}{$\boldsymbol{\alpha}$} &\multicolumn{3}{c}{\textbf{SC-based methods}} &\multicolumn{4}{c}{\textbf{KM-based methods}} &\multicolumn{3}{c}{\textbf{FCM-based methods}} &\multicolumn{3}{c}{\textbf{NMF-based methods}}\\
\cmidrule(r){2-4} \cmidrule(r){5-8} \cmidrule(r){9-11} \cmidrule(r){12-14}
& \textcolor{mygray}{\textbf{SC\_central}} & \textbf{FedSC} & \textbf{OmniFC-SC}
& \textcolor{mygray}{\textbf{KM\_central}} & \textbf{k-FED} & \textbf{MUFC} & \textbf{OmniFC-KM}
& \textcolor{mygray}{\textbf{FCM\_central}} & \textbf{FFCM} & \textbf{OmniFC-FCM}
& \textcolor{mygray}{\textbf{NMF\_central}} & \textbf{FedMAvg} & \textbf{OmniFC-NMF} \\
\hline
1000 & \textcolor{mygray}{\multirow{3}{*}{0.72}} & 0.71 & \textbf{0.72}
& \textcolor{mygray}{\multirow{3}{*}{0.62}} & 0.58 & 0.61 & \textbf{0.62}
& \textcolor{mygray}{\multirow{3}{*}{0.65}} & 0.60 & \textbf{0.66}
& \textcolor{mygray}{\multirow{3}{*}{0.70}} & 0.33 & \textbf{0.72} \\
5 &  & \textbf{0.73} & 0.72
&  & 0.48 & 0.61 & \textbf{0.62}
&  & 0.52 & \textbf{0.66}
&  & 0.33 & \textbf{0.72} \\
0.001 &  & 0.59 & \textbf{0.72}
&  & 0.51 & 0.60 & \textbf{0.62}
&  & 0.54 & \textbf{0.66}
&  & 0.33 & \textbf{0.72} \\
\hline
count & - & 1 & 2 & - & 0 & 0 & 3 & - & 0 & 3 & - & 0 & 3 \\
\hline\hline
\end{tabular}
}
\end{table}

\subsection{The Feasibility of Extending Deep Clustering}

To further showcase the flexibility of the proposed framework, we validated the feasibility of employing OmniFC to extend deep clustering. As a foundational deep k-means (KM) method, DCN \cite{yang2017towards} has spurred a range of advanced clustering techniques and found widespread use in privacy-sensitive domains such as medicine and finance \cite{miklautz2024breaking}. Therefore, extending DCN could significantly impact multiple research fields and accelerate progress on downstream applications.

To this end, we extend DCN analogously to KM by using the reconstructed distance matrix in place of raw features as the model input. This extended variant is denoted as OmniFC-DCN. As evidenced in Table \ref{tab:rebuttal_kappa}, OmniFC-DCN outperforms its shallow counterpart, OmniFC-KM, in clustering performance, further highlighting the flexibility of OmniFC.

\begin{table}[!t]
\centering
\caption{\textbf{Kappa of clustering methods in different simulated federated datasets.} OmniFC-DCN achieves superior clustering performance relative to its shallow counterpart, OmniFC-KM.}
\renewcommand{\arraystretch}{1.1}
\tabcolsep 10.25mm 
\begin{tabular}{lccc}
\hline\hline
Method & MNIST & Pendigits & 10x\_73k \\\hline
OmniFC-KM  & 0.42 & 0.62 & 0.56 \\
OmniFC-DCN & \textbf{0.43} & \textbf{0.64} & \textbf{0.60} \\
\hline\hline
\label{tab:rebuttal_kappa}
\end{tabular}
\end{table}

\section{Limitation}
\label{lim}

This work primarily focuses on extending shallow centralized clustering methods and may be less effective for high-dimensional or intrinsically complex data. A promising future direction is to explore how the reconstructed global distance matrix can substantially support the federated extension of deep centralized clustering methods, thereby enabling more powerful representation learning under complex data distributions \cite{zhou2024comprehensive, yan2022privacy, yan2024ccfc, yan2024ccfc++}.

\newpage

\end{document}